\documentclass{article}

\usepackage{xcolor}
\usepackage{multirow}
\usepackage{booktabs}
\usepackage{array}
\usepackage[table]{xcolor}
\usepackage{tabularx}
\usepackage{ragged2e}
\usepackage{makecell} 
\usepackage{float}
\usepackage{arxiv}

\usepackage[utf8]{inputenc} 
\usepackage[T1]{fontenc}    
\usepackage{hyperref}       
\usepackage{url}            
\usepackage{booktabs}       
\usepackage{amsfonts}       
\usepackage{nicefrac}       
\usepackage{microtype}      
\usepackage{lipsum}
\usepackage{fancyhdr}       
\usepackage{graphicx}       
\graphicspath{{media/}}     
  
\title{Efficient Vision-Language-Action Models for Embodied Manipulation: A Systematic Survey
}

\author{
\large
Weifan Guan$^{1,2}$, Qinghao Hu$^{1,\textsuperscript{†}}$, Aosheng Li$^{2,3,4}$, Jian Cheng$^{1,3\textsuperscript{†}}$\thanks{~\textsuperscript{†} Corresponding authors} \\[0.4em]
$^{1}$Institute of Automation, Chinese Academy of Sciences \\[-0.1em]
$^{2}$University of Chinese Academy of Sciences $~^{3}$AiRiA  \\[-0.1em]
$^{4}$Nanjing University of Information Science and Technology \\[-0.1em]
\\[-0.1em]
\texttt{guanweifan2024@ia.ac.cn},\;
\texttt{huqinghao2014@ia.ac.cn},\\
\texttt{liaosheng24@mails.ucas.ac.cn},\;
\texttt{jcheng@nlpr.ia.ac.cn}
}

\begin{document}
\maketitle

\begin{abstract}
Vision-Language-Action (VLA) models extend vision-language models to embodied control by mapping natural-language instructions and visual observations to robot actions. Despite their capabilities, VLA systems face significant challenges due to their massive computational and memory demands, which conflict with the constraints of edge platforms such as on-board mobile manipulators that require real-time performance. Addressing this tension has become a central focus of recent research. In light of the growing efforts toward more efficient and scalable VLA systems, this survey provides a systematic review of approaches for improving VLA efficiency, with an emphasis on reducing latency, memory footprint, and training and inference costs. We categorize existing solutions into four dimensions: model architecture, perception feature, action generation, and training/inference strategies, summarizing representative techniques within each category. Finally, we discuss future trends and open challenges, highlighting directions for advancing efficient embodied intelligence. The papers covered in this survey are compiled in a GitHub repository: \href{https://github.com/guanweifan/awesome-efficient-vla}{Awesome Efficient VLA}.

\end{abstract}


\section{Introduction}
Traditional robotic systems rely on task-specific algorithms and hand-designed rules, which perform well in structured environments but struggle to generalize to unstructured, real-world settings. The rise of deep learning has transformed this paradigm by enabling models to automatically extract hierarchical and transferable features. Building on this foundation, Vision-Language Models (VLMs) emerge as large pre-trained multimodal systems that align visual inputs with natural language, providing unified semantic representations across modalities. Extending this concept to embodied control, Vision-Language-Action (VLA) models establish an end-to-end mapping from language and vision to robot actions, enabling general-purpose control across diverse scenes and tasks. Compared to traditional methods, VLA model possesses superior semantic understanding and generalization capabilities, offering a new technological pathway for applications in robotics, including manipulation, navigation and beyond.

Despite their potential, VLA model faces substantial efficiency challenges that limit practical deployment. Many contemporary VLA system reuses large language models and heavy visual backbones, resulting in large parameter counts, high memory footprints, and slow inference. In addition, training these multimodal model is computationally expensive. These properties conflict with common robotic constraints, limited on-board compute, energy budgets and strict real-time latency requirements, thus impeding deployment on edge platforms such as mobile manipulators. While efficiency optimization has been extensively studied in VLM \cite{llmsurvey}, directly applying these techniques to VLA system is not straightforward. Because VLA introduces additional challenges: they must generate temporally consistent action sequences, operate under real-time constraints, and ensure physical reliability during execution. As a result, aggressive compression or pruning can easily degrade performance in ways that are far more critical than in VLM. These differences call for a dedicated examination of efficiency in VLA model, beyond what existing VLM-oriented methods can provide.

The rapid development of VLA has already inspired several surveys. For example, \cite{ma1778survey} provides a broad overview of concepts, architectures, training methods, and applications; \cite{zhong2025survey} analyzes different action representations in detail, comparing their respective strengths and weaknesses; \cite{jiang2025survey} focuses on autonomous driving as a specific application domain. However, none of these surveys have offered a systematic review from the perspective of efficiency, which has become a central bottleneck for real-world adoption as model sizes grow and real-time demands increase. This paper aims to fill this gap by presenting the first systematic survey dedicated to efficient VLA model.

Our contributions are threefold:

\begin{enumerate}
    \item To the best of our knowledge, this is the first survey dedicated to efficient VLAs, and we categorize efficiency-improvement techniques into four dimensions: model architecture, perception feature, action generation, and training/inference mechanisms.
    \item Based on this taxonomy, we summarize the mainstream approaches for enhancing VLA efficiency and analyze their strengths and weaknesses.
    \item We discuss future development trends of VLAs and highlight the aspects that should be prioritized to further improve efficiency under these trends.
\end{enumerate}

\begin{figure}[htbp]
  \centering
  \includegraphics[width=0.85\textwidth]{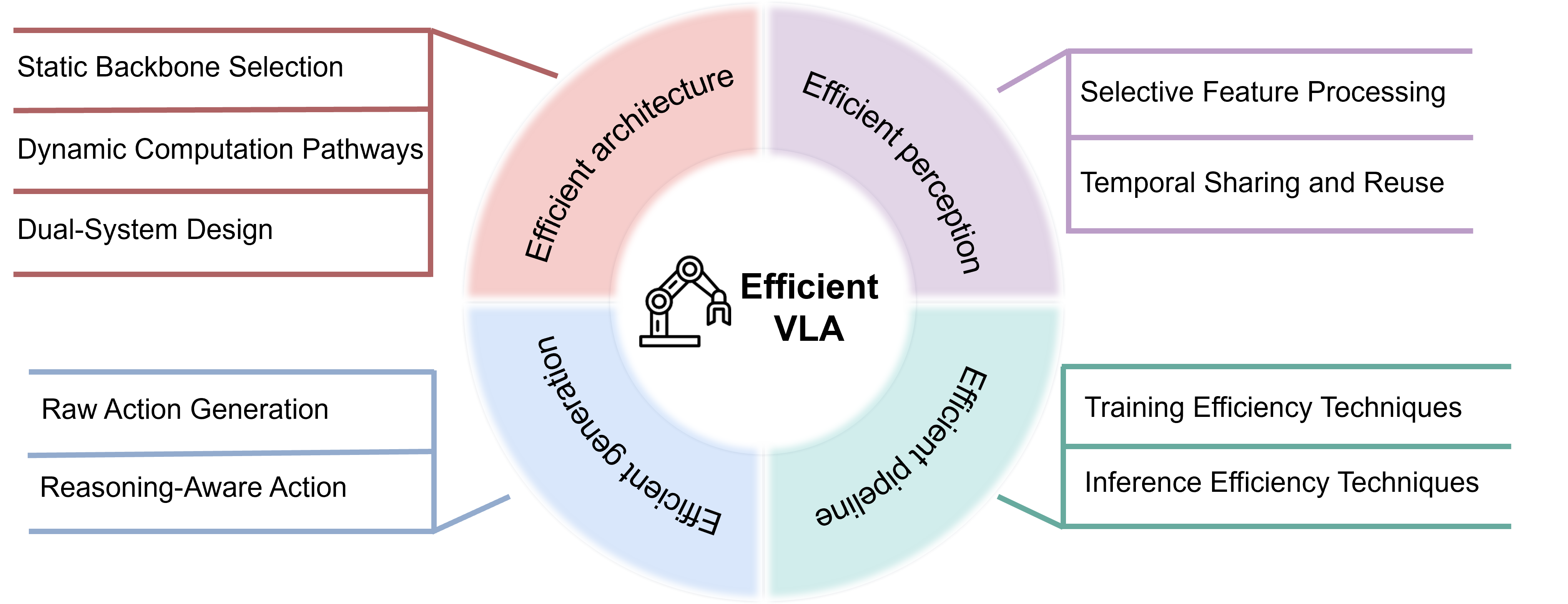} 
  \caption{Overview of the survey structure. The discussion on efficiency is organized into four core dimensions: efficient model architecture, efficient perception feature, efficient action generation, and efficient training-inference strategies.} 
  \label{1-1} 
\end{figure}

To realize these contributions, this survey follows the processing pipeline of VLA systems, as shown in Figure~\ref{1-1}, covering architectures, perceptual feature, action generation, and training/inference, and provides a focused review of techniques that improve efficiency, concluding with a forward-looking discussion of open challenges and promising directions. The remainder of this paper is organized as follows:

\textbf{- Section 2: Evolution of VLA Models.} We review the developmental trajectory of VLA models, tracing their evolution and summarizing the representative milestones and technical improvements that have shaped their current landscape.

\textbf{- Section 3: Efficient model architecture.} We examine how to build computationally efficient VLA architectures from two perspectives: static backbone selection and dynamic computation-path planning. It also discusses the potential efficiency gains of dual-system designs.
    
\textbf{- Section 4: Efficient perception feature.} We survey methods for reducing front-end cost, including removing spatial redundancy within single frame and reusing perception features across time steps.
    
\textbf{- Section 5: Efficient action generation.} We analyze and compares two mainstream action representations—raw actions and reasoning-based actions—and review methods to accelerate their generation.
    
\textbf{- Section 6: Efficient training and inference.} We review optimization techniques across the model lifecycle, covering cost-effective training paradigms and inference optimizations for deployment.

\textbf{- Section 7: Future prospects.} We address the future development of VLA models and outlines the necessary research foci for improving the efficiency of VLA to meet these emerging trends.

\section{Foundations and Evolution of VLA Models}

Vision-Language-Action models have recently emerged as an end-to-end paradigm for robot control. Their central idea is to design a unified model that directly maps high-dimensional perceptual inputs (Vision) and natural language instructions (Language) into low-level robot control signals (Action). By adopting this data-driven approach, VLAs aim to enhance generalization and semantic understanding, enabling robots to handle complex and unstructured tasks. The development of this line of research can be roughly divided into three stages.

\begin{figure}[htbp]
  \centering
  \includegraphics[width=0.8\textwidth]{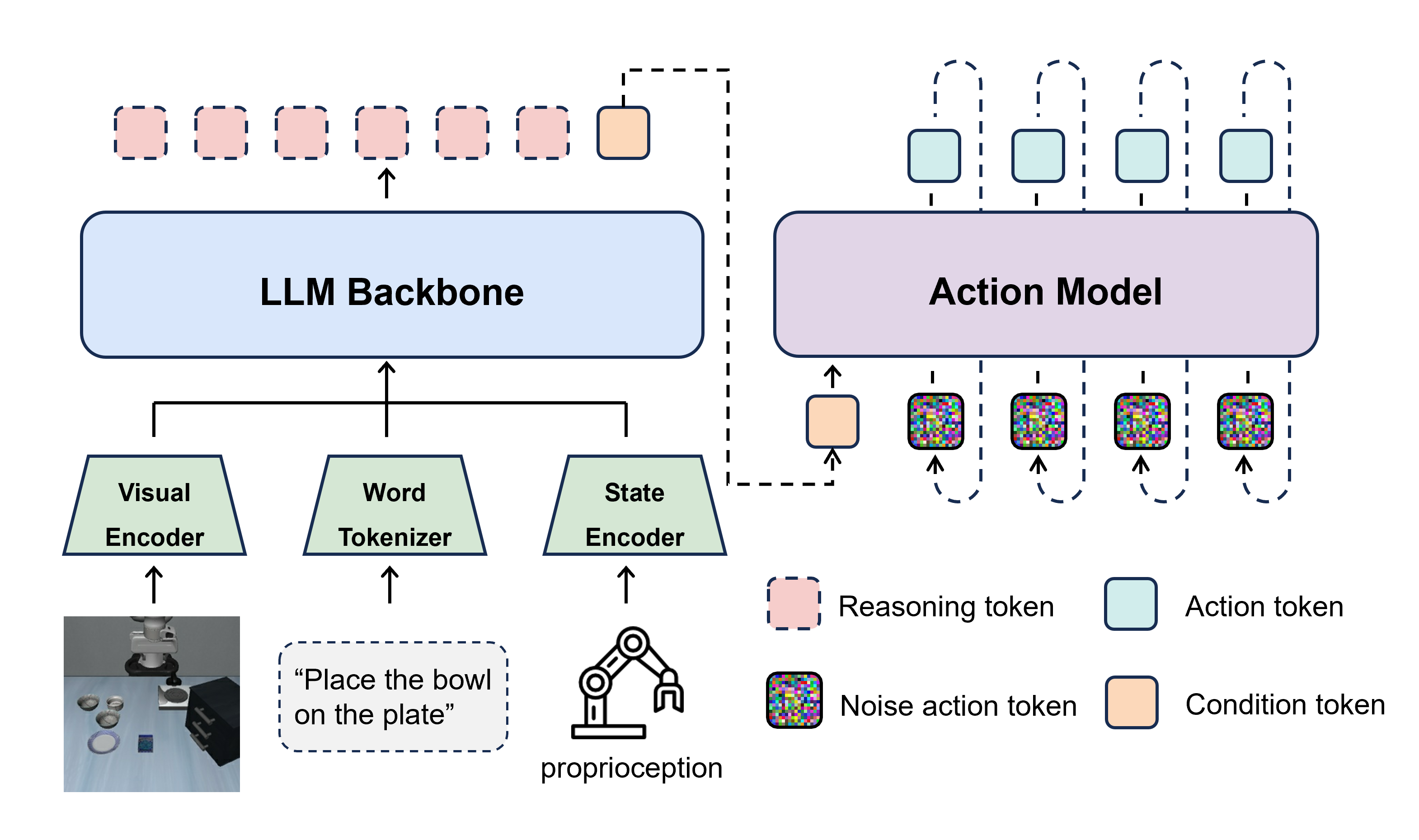} 
  \caption{Illustration of a typical VLA architecture. Visual observations, textual instructions, and robot proprioceptive states are first encoded and fused, then passed to an LLM backbone for reasoning. The resulting latent representation, enriched with planning information, is fed into a diffusion-based action model, where flow-matching optimization produces continuous action outputs.} 
  \label{2-1} 
\end{figure}

\textbf{Stage I: Early exploration and foundations.} Before the term VLA is formally established, researchers already begin to apply deep learning models directly to robot control. Works such as RT-1\cite{rt1} and Diffusion Policy\cite{diffusionpolicy} demonstrate the feasibility of using Transformers or diffusion models to build end-to-end policies. These models map raw RGB images and text instructions to action sequences. However, due to relatively small model sizes and limited datasets, their generalization remains constrained to specific tasks and environments, making it difficult to handle unseen instructions or objects beyond the training distribution.

\textbf{Stage II: Introduction of VLMs.} A turning point comes with the integration of pre-trained vision-language models (VLMs). RT-2\cite{rt2} formally introduces the concept of VLA and, for the first time, employs a pre-trained VLM as the backbone. By fine-tuning the VLM on large-scale robot trajectory datasets and converting robot control signals into discrete action tokens, RT-2 successfully transfers the general visual and semantic knowledge acquired from internet data to robot control tasks, achieving a significant leap in generalization. This "VLM-as-a-policy" paradigm quickly becomes mainstream. The release of OpenVLA\cite{openvla} further accelerates adoption by providing a standardized architecture: strong vision encoders such as SigLIP\cite{siglip} or DINO-v2\cite{dinov2} for visual feature extraction, a LLaMA tokenizer\cite{llama} for text instruction embedding, and a large language model such as LLaMA-7B for high-level reasoning. The LLM output is then used to predict discrete action tokens, which share the same vocabulary as language tokens. Building on this foundation, many follow-up works appear: Octo\cite{octo} explores diffusion models as policy heads for continuous action generation; the GR series \cite{gr1,gr2,gr3} and VPP\cite{vpp} adopt ideas from video generation to design pre-training tasks for VLA optimization; ConRFT\cite{conrft} investigates training paradigms that combine offline and online reinforcement learning. Meanwhile, the release of large and diverse real-world robot datasets such as Open X-Embodiment (OXE)\cite{oxe} and DROID\cite{droid} provides the data support needed to train increasingly large models.

\begin{figure}[htbp]
  \centering
  \includegraphics[width=0.8\textwidth]{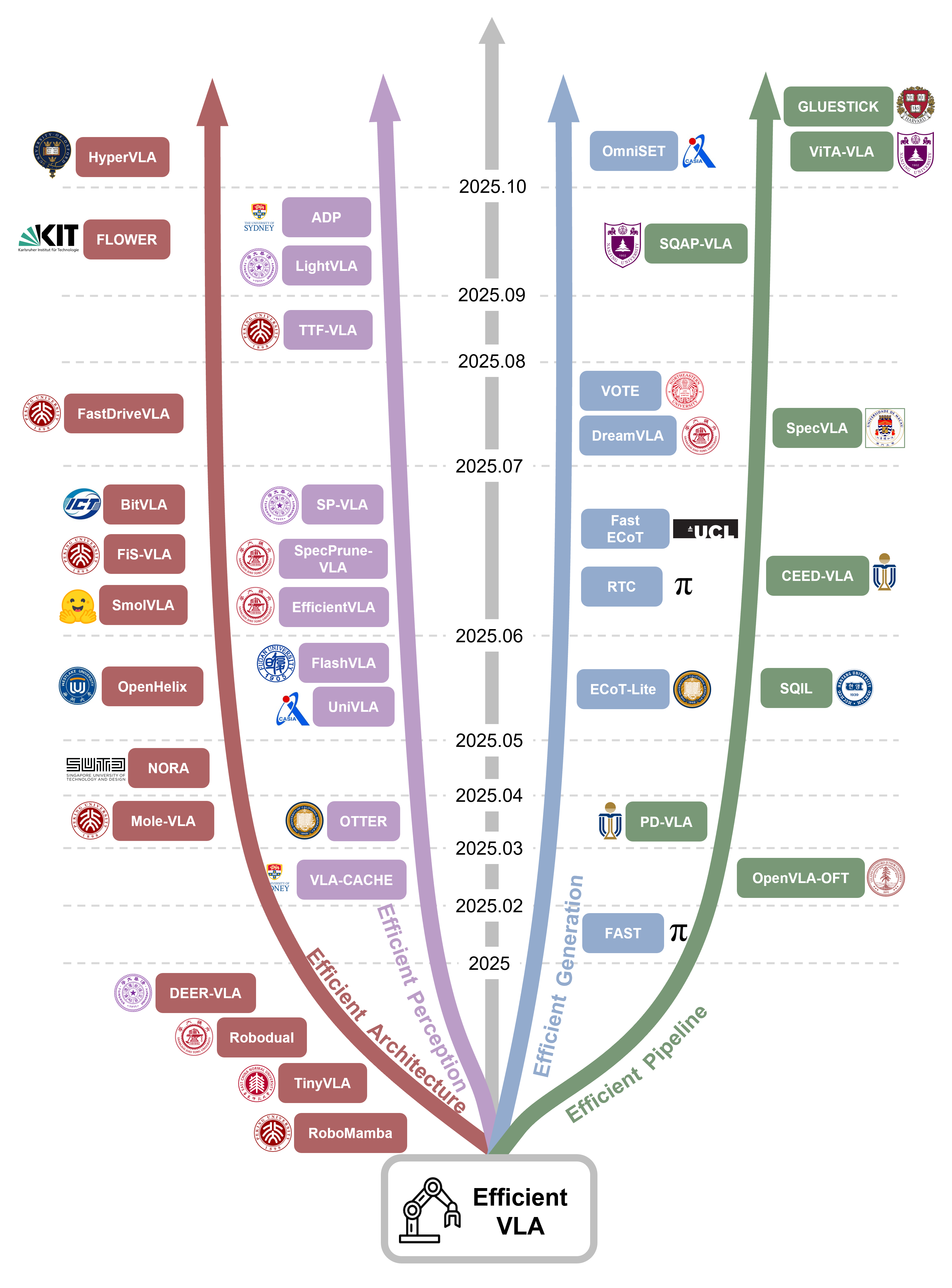} 
  \caption{Development trajectory of efficient VLA algorithms. The diagram highlights representative works from the past years that focus on improving the efficiency of VLA models.} 
  \label{2-2} 
\end{figure}

\textbf{Stage III: Architectural convergence and performance refinement.}  With further progress, VLA architectures begin to converge. A new generation of designs, exemplified by $\pi$0\cite{pi0}, becomes the prevailing choice. Typically, a pre-trained LLM handles high-level planning and intent understanding by integrating visual, language, and robot state information, and produces an abstract plan or policy representation. This is then passed to a dedicated action expert module, often based on diffusion transformers\cite{dit}, which refines the plan and generates smooth, precise continuous actions through mechanisms such as flow matching \cite{flowmatching}. An illustration of the current VLA architecture is shown in Figure~\ref{2-1}. However, in this stage, the models are generally large and inference remains slow. For example, OpenVLA has 7B parameters and runs at 5 Hz, while $\pi$0 contains 3B parameters and achieves around 10 Hz. These speeds are obtained even on powerful GPUs; on resource-constrained edge devices, such requirements in memory, computation, and latency will become more challenging to sustain. This motivates the recent surge of research on efficient VLA models, which aim to reduce parameter scale and accelerate inference. Their development trajectory is illustrated in Figure~\ref{2-2}. The following chapters will examine these developments in detail.

\section {Efficient Model Architectures}

Model architecture is the primary determinant of system efficiency: it directly affects training cost, inference latency, and storage requirements. Consequently, designing efficient foundational architectures remains a central focus of research. This section provides a systematic overview of principal approaches to efficient architectural design in Vision-Language-Action (VLA) models, organized into static backbones, dynamic computation pathways, and dual-system designs.

\subsection{Static Backbone Selection}
Contemporary VLA models typically rely on large-scale pretrained VLM backbones, leveraging broad world knowledge acquired during pretraining. Early work, in pursuit of broader generalization, often adopts very large VLMs with high parameter counts. While this trend improves task coverage, it also leads to substantial model size and heavy computational overhead in the first generation of VLA models. For example, RT-2\cite{rt2} reaches 55B parameters, which poses challenges for sequential and low-latency robotics tasks, as inference runs only at around 3 Hz. Empirical studies indicate that most latency comes from the language-model component, motivating the use of lighter LMs or efficiency-oriented LM designs as common strategies.

RoboMamba\cite{robomamba} introduces Mamba, a state-space model architecture, as its sequence model. At about 2.7B parameters, Mamba\cite{mamba} delivers more efficient temporal modeling and parallel inference than comparable Transformer-based LLMs, reducing latency with minimal loss in task performance. TinyVLA\cite{tinyvla} represents one of the early attempts toward VLA efficiency. By using smaller LMs such as Pythia-1.3B\cite{pythia}, it compresses the overall model while maintaining core task capabilities, making edge deployment more feasible. SmolVLA\cite{smolvla} takes a more direct structural simplification approach. It employs SmolVLM-2\cite{smolvlm} with parameter sizes of 0.24B, 0.45B, and 2.25B, and further reduces computation by pruning several of the final Transformer layers. Similarly, NORA\cite{nora} replaces the backbone with Qwen-2.5-VL-3B\cite{qwen}, yielding a smaller footprint while sustaining strong performance. 

Overall, these efforts share a common emphasis on downsizing—replacing large-scale backbones with lightweight alternatives to achieve efficiency gains without sacrificing essential task competence. This trend toward compact architectures has become increasingly prominent, as modern VLA systems continue to shift from multi-billion-parameter models to those with around one billion or even a few hundred million parameters, reflecting a broader movement toward practical and deployable embodied intelligence.


\subsection{Dynamic Computation Pathways}

The previous section highlighted the strategy of adopting more compact backbones to improve inference efficiency. While this approach is straightforward and effective, it inevitably narrows the model’s capacity ceiling. An alternative line of research instead retains large-scale backbones during training but introduces dynamic pathway selection during inference. In this way, the model preserves the expressive power of large architectures while discarding redundant computations in task-specific contexts, thereby achieving greater efficiency without fully compromising capability.

SmolVLA \cite{smolvla} adopts a simple layer-pruning strategy by permanently removing a fixed number of final layers from the language model. FLOWER \cite{flower} follows a semantically motivated pruning paradigm grounded in interpretability findings of large language models. It observes that intermediate Transformer layers capture rich, general semantics, whereas the final layers tend to over-specialize in next-token prediction. Accordingly, FLOWER prunes the redundant upper layers, removing the decoder in encoder-decoder VLMs and the last several layers in decoder-only ones, to balance contextual expressiveness and computational efficiency.

However, this form of pruning is static, as it neither adapts to the complexity of the input nor accounts for the specific demands of the task. To address this limitation, DEER-VLA \cite{deervla} incorporates an early-exit mechanism into the VLA system. It places lightweight policy heads at various intermediate layers of the language model, enabling action prediction at multiple depths. An output-similarity metric is then used to determine whether to exit early. The exit threshold is optimized via a constrained objective that balances average/peak FLOPs and GPU memory usage, rather than being tuned manually.

Nevertheless, MoLE-VLA \cite{molevla} argues that deep features remain critical for task performance, and thus premature termination risks undermining the model’s upper performance bound. To mitigate this, MoLE-VLA treats each layer of the language model as a potential expert and employs a Mixture-of-Experts (MoE) framework \cite{moe}. A gating mechanism dynamically selects which layers participate in the computation for a given input, thereby avoiding the complete loss of deeper-layer information. To stabilize training, self-distillation \cite{selfdistill} is further applied, where the full, unpruned network provides guidance to the reduced computational path.

Beyond MoE-based routing, another line of work leverages similarity-based skipping. Efficient-VLA \cite{efficientvla} evaluates the contribution of each layer by measuring the cosine similarity between its input and output feature vectors. If the similarity exceeds a threshold, indicating limited representational transformation, the layer is skipped during inference. In contrast to static pruning, this method adapts to input characteristics and preserves the potential availability of all layers, allowing the model to retain its full representational depth when needed while compressing computation in redundant cases.

\subsection{Dual-System Design}

\begin{figure}[htbp]
  \centering
  \includegraphics[width=0.85\textwidth]{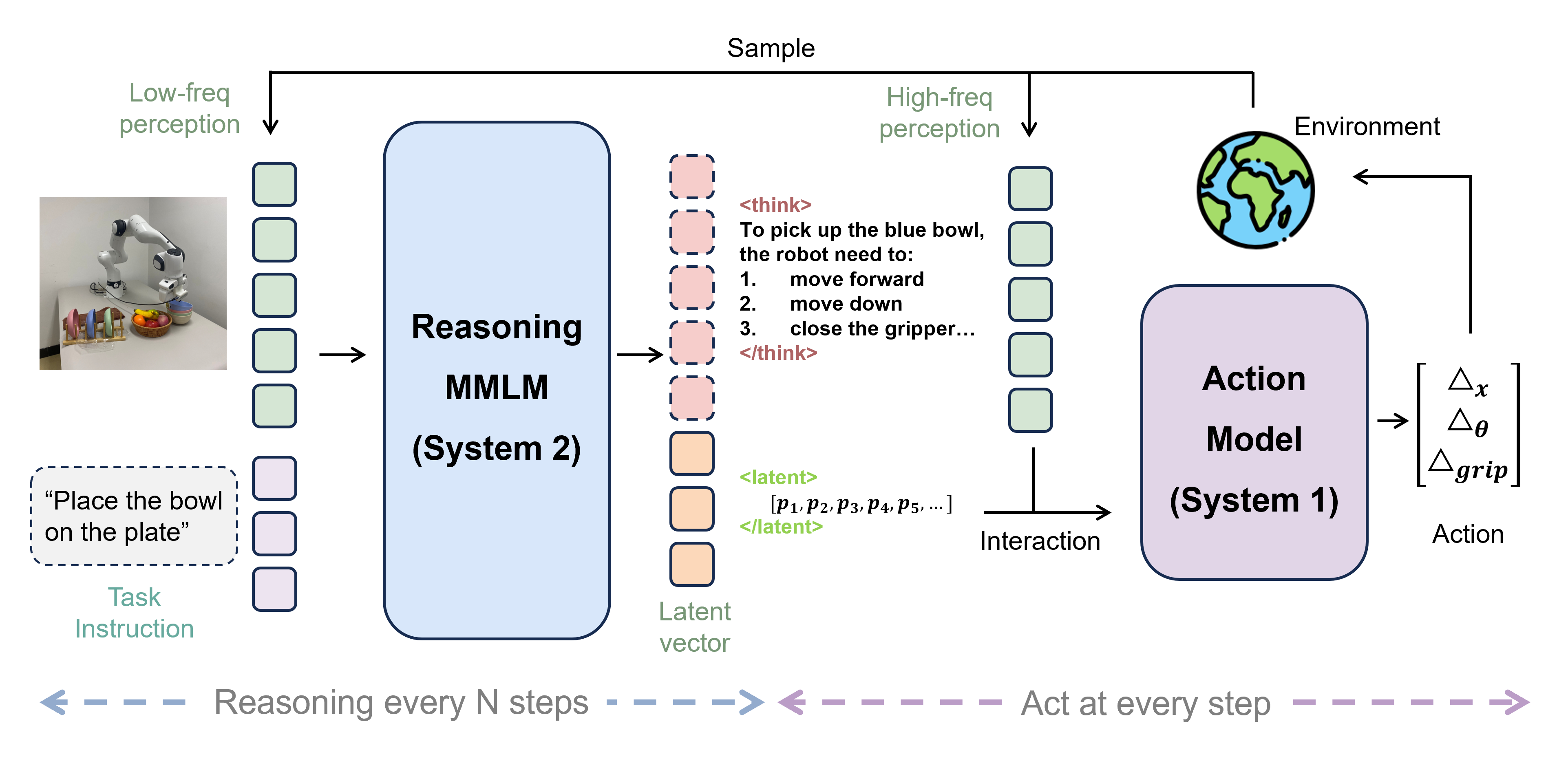} 
  \caption{Dual-system VLA framework. A multimodal LLM (System 2) processes infrequently updated visual and textual tokens, generating reasoning and latent tokens. The latent tokens are then consumed by a lightweight action model (System 1), which combines them with frequently updated visual tokens to produce raw actions. The two systems operate asynchronously: each round of System 2 reasoning provides latent vectors that enable multiple steps of System 1 inference.} 
  \label{3-1} 
\end{figure}

Beyond optimizations of backbones and inference pathways, another line of exploration lies in architectural reconfiguration through dual-system design. Inspired by dual-system theory in cognitive science \cite{thinking}, this approach divides the model into a slow system for complex reasoning and long-term planning, and a fast system for rapid, intuitive responses. Operating in concert, the two subsystems enable VLA models to manage intricate, high-level tasks while simultaneously ensuring low-latency inference in simpler scenarios. This strategy typically employs heterogeneous model architectures: the slow system relies on a large-scale multimodal language model (MMLM) to meet demands of semantic understanding and reasoning, while the fast system adopts a lightweight model to respond quickly to perceptual inputs. The two systems exchange information via latent tokens or embeddings to collaborate on task completion. A schematic overview of the dual-system VLA framework is shown in Figure~\ref{3-1}. Notable examples following this include:

\textbf{- LCB}\cite{lcb} employs LLaVA\cite{llava} as the slow system to generate language descriptions and action prompts, which then guide a 3D Diffuser Actor\cite{3ddiffueractor} as the fast system to produce the final action through a learnable <ACT> token.
    
\textbf{- HiRT}\cite{hirt} adopts InstructBLIP\cite{instructblip} as the slow system to generate representations, which are subsequently processed by an EfficientNet-B3\cite{efficientnet} fast system via MAP pooling \cite{map} for efficient control.
    
\textbf{- RoboDual}\cite{robodual} combines OpenVLA\cite{openvla} as the slow system with a DiT\cite{dit} as the fast system. The slow system outputs latent representations, which the fast system refines through a Perceiver Resampler \cite{resampler} to reconstruct simplified action outputs.
    
\textbf{- OpenHelix}\cite{openhelix} provides a systematic review and evaluation of mainstream dual-system frameworks and proposes an optimized modular configuration. Specifically, LLaVA-7B\cite{llava} serves as the slow system and the 3D Diffuser Actor\cite{3ddiffueractor} as the fast system, with a learnable <ACT> token appended to the input sequence. The output of this token conditions the Actor. During training, the slow system is optimized with an auxiliary objective of reconstructing 3D positions and rotations, while the fast system continues to be trained with action denoising as its primary task.
    
\textbf{- FiS}\cite{fis} integrates the fast and slow systems into a single network, forming an implicit dual-system pathway. Shallow layers construct intermediate semantic representations, which are then consumed by the final layer to predict actions.

\textbf{- Hume}\cite{hume} introduces a cascaded dual-system structure. The slow system generates candidate action chunks at multiple noise scales, while a learnable aggregation token feeds into a value query head that scores the candidates. The most promising chunk is then further decomposed and denoised by the fast system to yield the final action sequence. Training is conducted jointly: the policy head and fast system are optimized via flow matching\cite{flowmatching}, while the value query head is trained with offline reinforcement learning on reward-annotated datasets.

Recent research has explored non-standard dual-system architectures for Vision-Language-Action (VLA) models, where two complementary subsystems interact in nontraditional ways. HyperVLA\cite{hypervla}, for instance, proposes a two-stage framework in which a HyperNetwork dynamically generates task-specific Base Policy parameters conditioned on language instructions and visual inputs. The compact Base Policy then performs single-step action generation using visual features and a learnable action token, thus eliminating the need for language instructions and reducing the computational cost of autoregressive or diffusion-based decoding. Conceptually, HyperVLA can be viewed as a parameter-transmitting dual-system VLA: the HyperNetwork (System 2) encodes perception into the parameters of the Base Policy (System 1), enabling efficient and semantically consistent action generation.

It is worth noting that most of the aforementioned methods still adopt a serial architecture, where the fast system’s response depends on prior processing by the slow system. To address this bottleneck, SP-VLA \cite{spvla} distinguishes between intuitive and deliberate actions based on end-effector velocity. Intuitive actions are handled by a lightweight model (e.g., ridge regression\cite{ridge}), whereas deliberate actions are processed by the main VLA model. The lightweight pathway is activated only when the proportion of actions requiring full VLA-level processing exceeds a predefined threshold, ensuring both responsiveness and reliability.
    
A summary of these representative dual-system VLA algorithms is provided in Table~\ref{tab:dual_system_vlas}. Moreover, while not a traditional VLA architecture, RT-Cache \cite{rtcache} offers a complementary fast pathway grounded in experience replay. Visual features are extracted by SigLIP\cite{siglip} and DINO-V2\cite{dinov2}, followed by two-stage clustering and nearest-neighbor search to efficiently retrieve historical trajectories. This allows the system to rapidly generate actions in repetitive or low-variability environments. From a cognitive perspective, such experience-based retrieval can be regarded as an alternative instantiation of the fast system, closely mirroring how humans draw upon prior experience for rapid decision-making.

\begin{table}[h]
\centering
\scriptsize
\renewcommand{\arraystretch}{1.5}
\caption{Representative dual-system Vision-Language-Action models.}
\label{tab:dual_system_vlas}
\begin{tabularx}{0.7\textwidth}{
    >{\hspace{0.2cm}\raggedright\arraybackslash}m{2.4cm}  
    >{\raggedright\arraybackslash}m{2.7cm}         
    >{\raggedright\arraybackslash}m{2.7cm}         
    >{\raggedright\arraybackslash}X  
}
\toprule
\textbf{\centering VLA} & \textbf{\centering System 1 (Intuitive)} & \textbf{\centering System 2 (Deliberative)} & \textbf{\centering Communication} \\
\midrule
\addlinespace[2pt]
LCB\cite{lcb}       & 3D Diffuser Actor & LLaVA                & Special token \\
\addlinespace[2pt]
HiRT\cite{hirt}      & EfficientNet-B3   & InstructBLIP         & Latent vector   \\
\addlinespace[2pt]
RoboDual\cite{robodual}  & DiT               & OpenVLA              & Latent vector   \\
\addlinespace[2pt]
OpenHelix\cite{openhelix} & 3D Diffuser Actor & LLaVA-7B             & Special token \\
\addlinespace[3pt]
FiS\cite{fis}       & \makecell[lc]{LLaMA2-7B \\ (Deep-layer)} & \makecell[lc]{LLaMA2-7B \\ (Shallow-layer)} & Latent vector   \\
\addlinespace[3pt]
Hume\cite{hume}      & Transformer       & Paligemma-4B         & Noisy action  \\
\addlinespace[2pt]
HyperVLA\cite{hypervla}    & Transformer  & T5             & Network parameter            \\
\addlinespace[2pt]
SP-VLA\cite{spvla}    & Ridge Regression  & LLaMA-7B             & Adaptive routing            \\
\addlinespace[2pt]
\bottomrule
\end{tabularx}
\end{table}

\subsection{Summary and Analysis}

In this section, we reviewed three major strategies for efficient architectural design in VLA models: static model backbones, dynamic computation pathways, and dual-system architectures. Static backbones deliver direct efficiency gains by employing more compact models; dynamic computation pathways enable flexible routing during inference, balancing capacity and cost; and dual-system architectures, inspired by cognitive theory, distribute reasoning and reactivity across distinct subsystems to achieve hierarchical collaboration. Collectively, these approaches outline the current landscape of efficient VLA architectures.

However, each strategy carries notable limitations. Static backbones, when overly compressed, reduce the model’s capacity ceiling and compromise generalization, yielding performance that may appear strong on current tasks but does not transfer well to novel settings; dynamic computation pathways, though flexible, often require additional branching modules and substantial training overhead and careful manual design of selection criteria and thresholds; dual-system architectures, while effective in balancing complex reasoning and fast responses, are frequently implemented in asynchronous forms, which introduce delays between the outputs of the two subsystems and thereby undermine real-time decision-making. These challenges underscore the difficulty of striking an ideal balance between efficiency and capability.

Future research must advance in several directions. First, scaling laws tailored to VLA models should be explored by evaluating across diverse tasks and experimental conditions, in order to clarify trade-offs among model size, generalization, and efficiency, and to identify backbone scales best suited to current data availability. Second, dynamic computation pathways could benefit from adaptive and automated mechanisms, such as reinforcement learning for layer-skipping, so that the number of executed layers is determined online rather than fixed by manually designed heuristics. Finally, since many VLA use cases demand edge deployment, architectural designs should explicitly consider cloud-edge partitioning: lightweight fast subsystems can be deployed locally to ensure low-latency control, while heavier reasoning modules are run in the cloud. Such frameworks must account for communication latency, bandwidth constraints, and privacy requirements to ensure robust operation.

\section {Efficient Perception Feature}

 VLA systems typically operate on multi-modal inputs, including visual observation, textual instruction, and robot proprioception. Among these, visual inputs contribute the most to the total token sequence length, often dominating the model’s memory and compute footprint. Thus, dense processing of visual tokens is a primary source of computational burden and a primary target for efficiency optimization in VLA models.

However, not all visual information is equally relevant to the decision-making process. In many tasks, substantial portions of the input image, such as background regions, task-irrelevant objects, or temporally invariant content, do not significantly influence action selection. These regions generate redundant computation, particularly with high-resolution inputs and long-horizon tasks, thus reducing inference efficiency. To address this issue, a growing line of research has emerged, aiming to construct compact yet task-relevant visual representations. These efforts generally fall into two complementary directions:  

\textbf{(1) Selective processing of single-frame perception} : pruning, compressing or transforming redundant information before it reaches downstream policy networks;

\textbf{(2) Temporal sharing and reuse} : exploiting inter-frame similarity to avoid recomputation of static or slowly changing features.

\subsection{Selective Feature Processing}

A major efficiency bottleneck in deploying VLA models lies in their dense processing of high-dimensional visual inputs. As a result, a substantial body of work has focused on compressing perceptual representations, typically by reducing the length of token sequences input to downstream policy networks, while preserving task-relevant information.

\begin{figure}[htbp]
  \centering
  \includegraphics[width=0.85\textwidth]{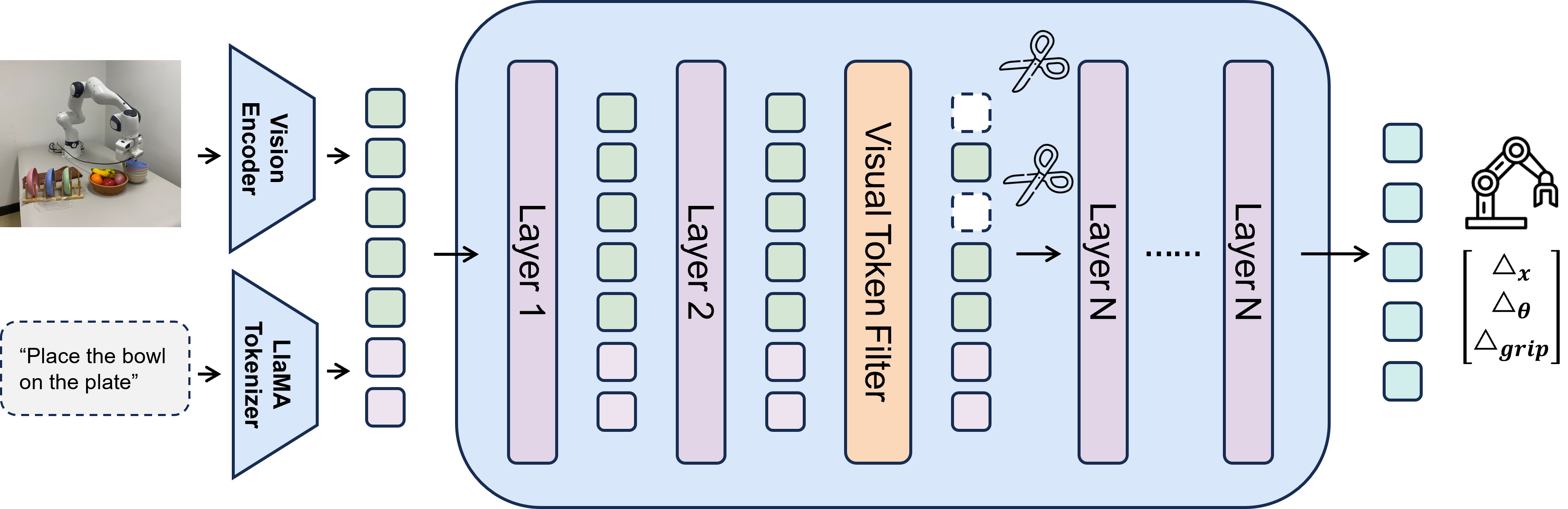} 
  \caption{Token pruning in VLA systems. During forward inference, visual tokens are scored based on importance metrics, and less informative tokens are pruned to reduce computation. This pruning can occur either before entering the LLM backbone or within its internal layers.} 
  \label{4-1} 
\end{figure}

A straightforward method is to prune tokens during inference. The schematic diagram is shown in Figure~\ref{4-1}. For example, FastV \cite{fastv} computes the average attention each visual token receives from all tokens at an intermediate LLM layer and applies Top-K pruning based on these importance scores. Building on this line, EfficientVLA \cite{efficientvla} further quantifies the interaction between visual tokens and task instructions, selecting key tokens that capture semantic relevance and augmenting them with task-driven tokens of high attention and diversity-driven tokens to ensure representational richness. SP-VLA \cite{spvla} emphasizes that token pruning should also preserve spatial structure; in addition to semantic importance captured through attention, spatial relevance is measured via contour cues using edge detection, and tokens satisfying either criterion are retained, ensuring scene integrity. SP-VLA also introduces adaptive pruning that modulates aggressiveness according to motion dynamics, trading efficiency for fidelity as task requirements vary. 

While simple and effective, these attention-score-based strategy relies on intermediate attention scores, which efficient implementations such as FlashAttention \cite{flashattention} do not expose. To address this limitation, FlashVLA \cite{flashvla} proposes a feature-based alternative by directly applying singular value decomposition (SVD) on attention output matrix to derive an Information Contribution Score (ICS), which measures each token’s projection onto dominant singular directions.

Dynamic and context-aware pruning strategies extend these static or attention-based approaches by incorporating language instructions and action information. LightVLA \cite{lightvla} targets the visual tokens produced by the vision encoder rather than operating within the LLM itself. It adopts a query-driven token selection mechanism, where queries dynamically generated through cross-modal attention identify the most informative visual tokens. The selection process is made differentiable using Gumbel-Softmax combined with a straight-through estimator, enabling end-to-end training while preserving spatial position encoding without the need to manually predefine how many tokens should be retained. ADP \cite{adp} introduces a two-stage mechanism: first, task-driven static pruning computes textual queries and cross-modal attention to evaluate the global importance of each visual token, retaining those most relevant to the instruction; second, a dynamic, action-aware switch adjusts pruning based on recent end-effector motion, using a hysteresis mechanism to balance compression during coarse movements and perceptual fidelity during precise operations. FASTDriveVLA \cite{fastdrivevla} similarly integrates action-awareness into token pruning but adds a foreground-background adversarial reconstruction during training, ensuring the model can distinguish critical foreground from redundant background information. Token scores are computed with a lightweight scorer that combines token features and a learnable query via Hadamard fusion, and Top-K selection is applied during inference while preserving positional information. SpecPrune-VLA \cite{specprunevla} is a training-free pruning method that performs two-level token reduction with heuristic control. At the action level, static pruning evaluates global token redundancy from previous actions and local token relevance for the current action, reducing visual tokens before generation. At the layer level, dynamic pruning exploits the relevance between tokens and model layers, pruning tokens based on layer-specific importance. A lightweight, action-aware controller further adjusts pruning according to the granularity of the current action—coarse-grained actions tolerate more pruning, while fine-grained actions require higher fidelity. 

Beyond dynamic pruning, maintaining robustness under low-precision quantization and ensuring spatial coverage are critical for practical deployment. SQAP-VLA \cite{sqapvla} addresses these challenges through a spatially and quantization-aware token pruning framework, combining three complementary mechanisms: preserving task-critical tokens under quantization, protecting tokens near the robot end-effector, and sampling tokens to maintain spatial coverage. These strategies together ensure that the final set of retained tokens balances efficiency, stability, and coverage, enabling reliable low-bit inference. The summary of different types of toekn-pruning VLA strategies is presented in the Table~\ref{tab:efficient_vla_methods}.

\begin{table}[h]
\begin{center}
\scriptsize
\renewcommand{\arraystretch}{2.0}
\caption{Representative token-pruning strategies in Vision-Language-Action models.}
\label{tab:efficient_vla_methods}
\begin{tabularx}{0.95\textwidth}{
    >{\raggedright\arraybackslash}m{2.2cm}  
    >{\raggedright\arraybackslash}m{4cm}  
    >{\raggedright\arraybackslash}m{4cm}  
    >{\raggedright\arraybackslash}m{4cm}  
}
\toprule
\textbf{~~~Method} & \textbf{~~~~~Token Importance Criterion} & \textbf{~~~~~~~~~~~~~~~~~~~~~~~~~~~~Pros} & \textbf{~~~~~~~~~~~~~~~~~~~~~~~~~~~~Cons} \\
\midrule
FastV \cite{fastv} & Top-K attention scores based on the average received attention of visual tokens at the intermediate LLM layer & Simple and effective; easy to implement & Requires intermediate attention; ignores spatial structure and action dynamics \\
EfficientVLA \cite{efficientvla} & Attention scores weighted by task instruction relevance and token diversity & Preserves task-relevant information and ensures representation richness & Static selection; limited adaptation to dynamic actions \\
SP-VLA \cite{spvla} & Attention-based semantic relevance and contour edge-based spatial importance & Maintains spatial integrity and semantic coverage; adaptive pruning adjusts to motion & Extra spatial computation; increased implementation complexity \\
FlashVLA \cite{flashvla} & Singular value decomposition of attention output to compute Information Contribution Score & Compatible with high-efficiency attention; does not require intermediate attention access & Static pruning; SVD computation adds overhead \\
LightVLA \cite{lightvla} & Cross-modal attention queries with Gumbel-Softmax differentiable selection & End-to-end trainable; preserves spatial positions; automatically determines number of tokens to retain & Training is complex; requires differentiable selection mechanism \\
ADP \cite{adp} & Task-driven attention scoring combined with action-aware dynamic adjustment using motion hysteresis & Balances compression and perceptual fidelity; adapts to end-effector motion & Implementation complexity; hysteresis parameters need tuning \\
FASTDriveVLA \cite{fastdrivevla} & Lightweight scorer combining token features and learnable query; Top-K selection with positional preservation & Preserves critical foreground tokens; dynamic and action-aware; efficient scoring & Training requires adversarial objective; sensitive to background variations \\
SpecPrune-VLA \cite{specprunevla} & Static action-level redundancy and local relevance combined with layer-level token importance & Training-free; fast deployment; adapts to coarse and fine actions & Heuristic rules may limit generalization; less optimal for complex tasks \\
SQAP-VLA \cite{sqapvla} & Preservation of task-critical tokens under quantization, robot-proximity protection, and spatial sampling & Robust low-bit inference; maintains spatial coverage and important tokens & Static pruning; less responsive to dynamic actions; more complex to implement \\
\bottomrule
\end{tabularx}
\end{center}
\end{table}

In contrast to the above methods that subtract from the original token set, another line explores more fundamental representational transformations, aiming to improve efficiency via compression or unification. For instance, OTTER\cite{otter} draws inspiration from the Perceiver \cite{resampler} framework and proposes a cross-attention pooling mechanism. Guided by text instructions, this module compresses visual and textual tokens into fixed-length, compact representations. The emphasis on compression rather than discarding offers a new paradigm for handling long input sequences. UniVLA\cite{univla} seeks to unify visual, textual, and action modalities by converting all inputs into discrete tokens drawn from a shared vocabulary. This homogeneous representation enables seamless integration across modalities and simplifies the training of downstream tasks such as Perception Grounding, World Modeling, and Policy Learning, thereby improving multi-modal integration. While tokenization reduces visual granularity, it can substantially lower sequence length and ease multi-task training.

\subsection{Temporal Sharing and Reuse}

Having discussed strategies for optimizing single-frame inputs, we now focus on a complementary axis, exploiting temporal redundancy in sequential decision-making. In robotic task execution, adjacent states often exhibit strong temporal correlation, making full re-computation at each timestep unnecessarily costly. A promising direction is thus to identify and reuse computation results that remain stable across time. This requires precise judgment of which components can be reused and which require updating, spanning from low-level perception to high-level reasoning.

At the visual representation level, VLA-cache\cite{vla-cache} exemplifies such reuse. It targets redundancy between consecutive frames by reusing the key-value (KV) caches of static image patches in Transformer-based architectures. These KV caches store Transformer key-value vectors, allowing the model to skip recomputation for unchanged patches. VLA-cache estimates patch-level similarity across frames and reuses KV-caches for patches deemed static. To avoid performance degradation, highly task-relevant tokens are excluded from reuse and freshly recomputed. Moreover, VLA-cache introduces attention entropy as a measure of confidence, dynamically adjusting the reuse ratio across layers to balance precision and efficiency.

Complementary to KV-cache strategies, TTF-VLA \cite{ttfvla} leverages temporal redundancy by selectively fusing visual tokens across consecutive frames. Instead of reusing Transformer states, it maintains a history of patch tokens and updates them via a binary importance mask that identifies regions with significant visual or semantic changes. The mask is derived from both pixel-level differences and attention-based relevance, ensuring that only dynamic or task-critical patches are recomputed. A periodic keyframe update further prevents long-term drift, enabling efficient temporal fusion with minimal accuracy loss.

At a more abstract level, researchers extend reuse to high-level perceptual representations that drive task execution. FlashVLA\cite{flashvla} observes that in many scenarios, especially when the environment remains stable, the internal representations that guide action selection change only minimally over time. It therefore introduces a lightweight trigger that detects both similarity in consecutive perception-driven states and overlap in selected visual tokens. When these conditions are met, the model reuses previously computed representations, avoiding redundant recalculation while ensuring consistency.

Temporal redundancy also appears within the iterative process of action reasoning. Diffusion-based architectures, such as DiT\cite{dit}, generate multi-step action representations through repeated denoising. Since intermediate features across iterations often remain similar, EfficientVLA\cite{efficientvla} adopts a fixed-interval caching strategy: it recomputes features every N steps and reuses cached representations in between. This mechanism leverages temporal locality in the generative process itself rather than across physical frames, substantially lowering computational overhead while maintaining action coherence.

The reuse principle also applies to reasoning-intensive VLA models that employ Chain-of-Thought (CoT)\cite{cot} prompting. ECoT\cite{ecot} has vision-language models first generate natural language subtask plans before producing task actions, which improves interpretability but increases inference latency. Fast ECoT\cite{fastecot} identifies that such high-level reasoning modules, especially planning, evolve slowly and exhibit strong temporal locality. It proposes module-level caching, where the system reuses the prior text-based plan if no significant changes are detected. By caching costly but slowly varying reasoning results and combining them with modern accelerators such as continuous batching, Fast ECoT achieves substantial efficiency gains without sacrificing accuracy.

\begin{figure}[htbp]
  \centering
  \includegraphics[width=0.9\textwidth]{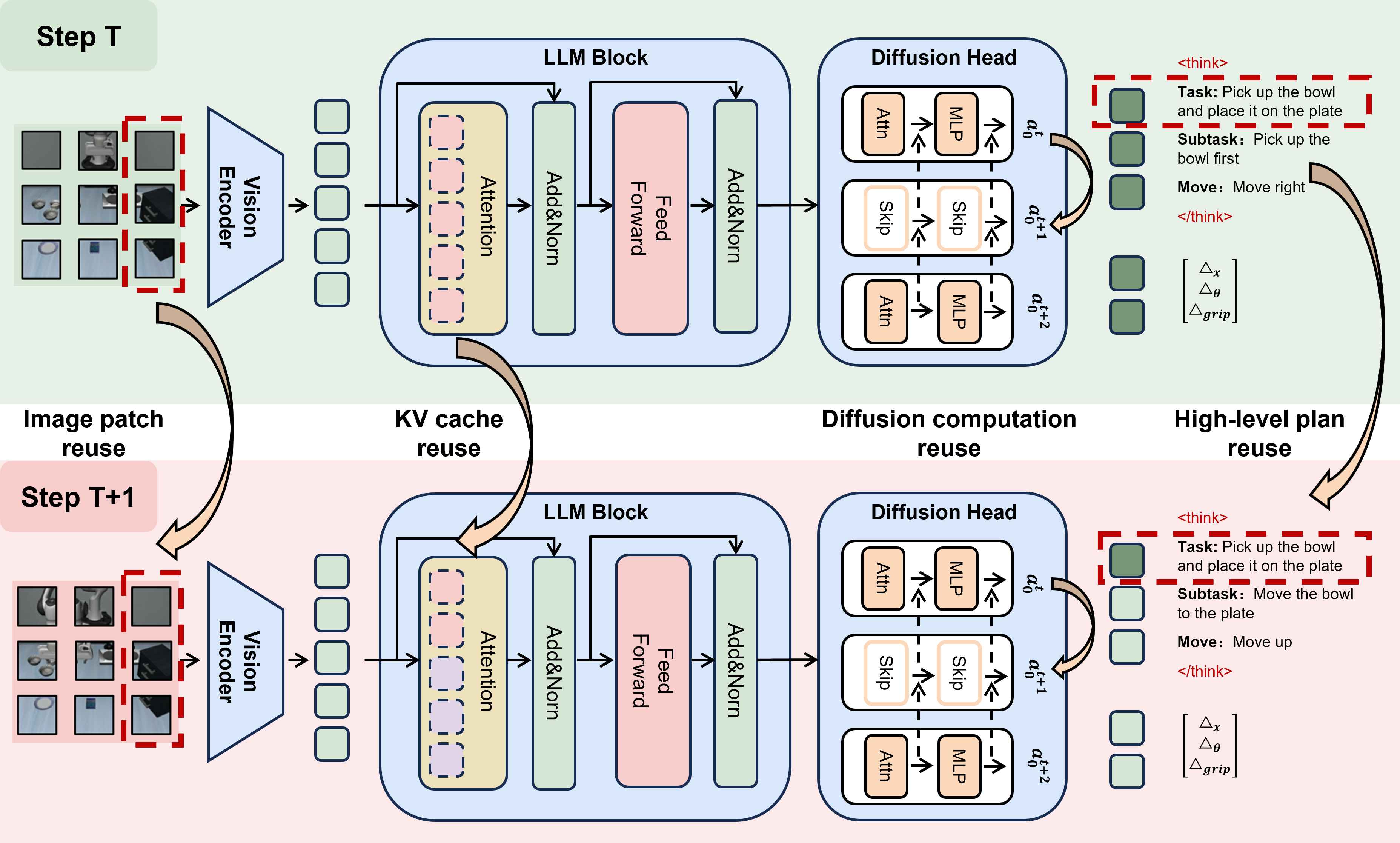} 
  \caption{Temporal reuse mechanisms in VLA systems. Existing approaches exploit similarities across timesteps, including patch reuse, KV-cache reuse, and high-level reasoning reuse, as these representations often change slowly over time. In diffusion-based action models, temporal reuse also arises naturally within the iterative denoising process.} 
  \label{4-2} 
\end{figure}

A summary illustration of temporal reuse in the VLA system pipeline is shown in the Figure~\ref{4-2}

\subsection{Summary and Analysis}

In this section, we systematically examined the impact of perceptual information representation on the efficiency of VLA systems from two complementary perspectives. On the one hand, redundant information in single-frame visual inputs can be pruned by evaluating the importance of tokens with designed indicators, thereby enabling feature compression and refinement. On the other hand, reusing cross-frame temporal features, intermediate computational results, and intermediate reasoning outputs reduces redundant processing and improves overall efficiency. Together, these approaches promote the streamlining of perceptual information, allowing VLA systems to achieve higher efficiency in reasoning and execution while maintaining perceptual capacity.

Despite these advances, current research still faces notable challenges. Token pruning methods often rely on manually predefined thresholds or fixed pruning ratios, lacking the ability to adapt dynamically to changing tasks or environments. As a result, a pruning strategy that performs optimally in one setting may degrade performance or even cause task failure in another. Furthermore, most perceptual representations remain limited to 2D RGB inputs. Recent efforts incorporate 3D representations to improve spatial awareness. However, 3D processing often incurs large computational overhead, compromising real-time performance and thus limiting practical use.

Looking ahead, future research is expected to move forward along two directions. First, the development of adaptive mechanisms for dynamically adjusting pruning ratios and strategies will be crucial, enabling models to optimize feature selection in response to task complexity and environmental variability and thereby achieving a more robust balance between efficiency and performance. Second, for 3D visual representations, progress will depend on more efficient modeling and compression strategies. Promising directions include lightweight depth estimation from monocular cues, voxel or point-cloud compression for reducing data density, and hybrid 2D-3D fusion methods that leverage the efficiency of image-based features while retaining essential spatial structure. Such approaches aim to preserve the benefits of 3D perception, such as depth reasoning and spatial consistency, without incurring prohibitive costs.

\section {Efficient Action Generation}

Compared to vision-language models, the key structural distinction of VLA model lies in the action as a pivotal component. Action serves as the closed-loop interaction interface between the model and the physical world, functioning as the bridge that transforms high-level semantic reasoning into low-level control commands. Consequently, the choice of action representation and generation strategy strongly affects control precision, response latency, and overall execution efficiency in long-horizon complex tasks.

\subsection{Raw Action Generation}

In mainstream VLA systems, the common design directly outputs a raw action in the form of a 7D vector, consisting of 3D translational offsets, 3D rotational offsets, and a binary gripper state. This low-dimensional continuous representation enables fast end-to-end control with minimal inference delay. However, when generating long action sequences, this step-by-step prediction scheme not only suffers from compound error that causes the executed trajectory to drift from the intended path, but also faces efficiency limitations. Because VLA input usually contains a long token sequence, if one input token corresponds to only one short action token, the overall action generation efficiency is low and inference speed becomes a bottleneck.

\begin{figure}[htbp]
  \centering
  \includegraphics[width=0.75\textwidth]{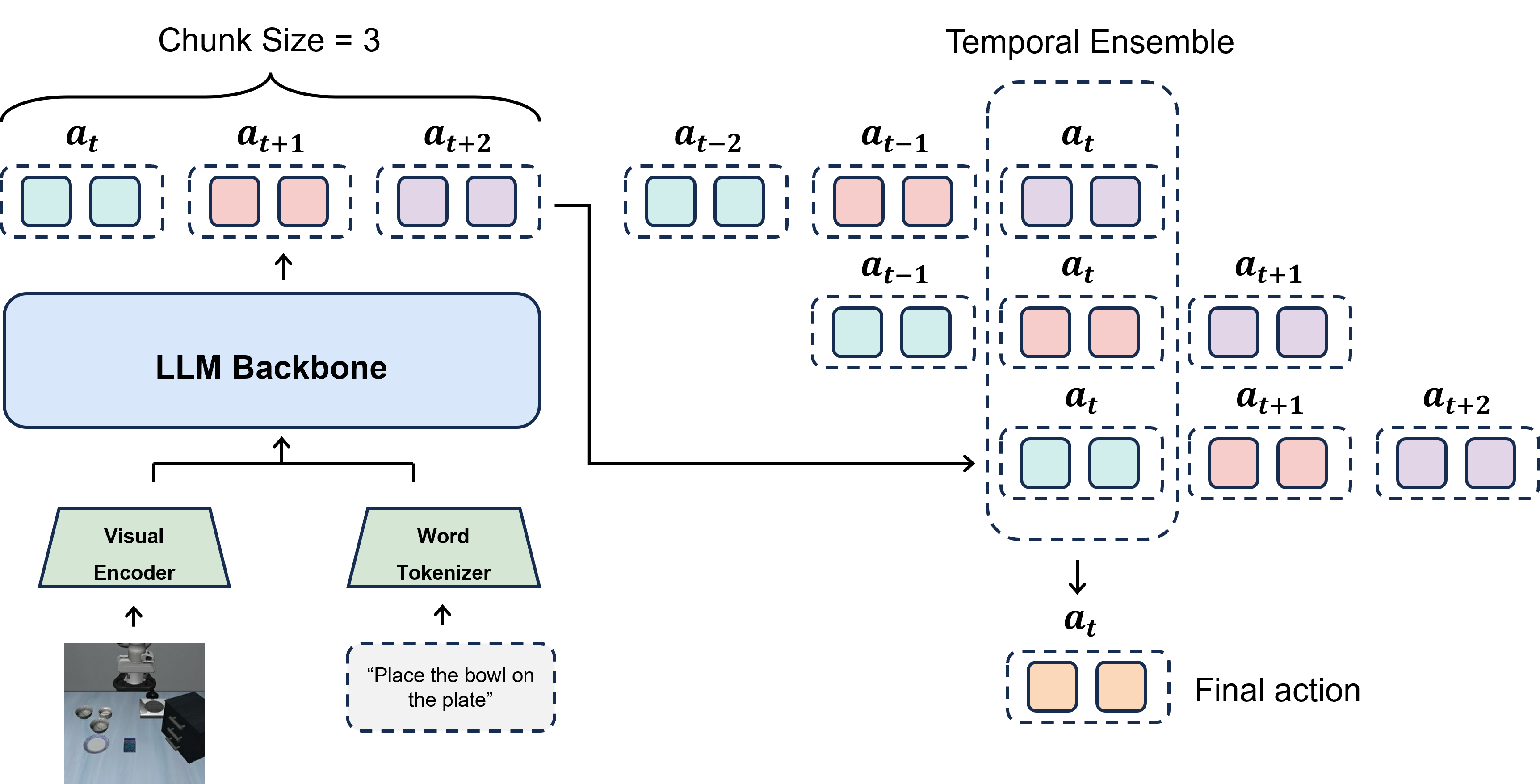} 
  \caption{Illustration of the action chunking mechanism. A single forward inference generates actions for multiple timesteps to improve efficiency, while temporal ensembling integrates them across time to reduce bias and variance.} 
  \label{5-1} 
\end{figure}

To mitigate this issue, Action Chunk\cite{act} generates a block of consecutive actions in one inference step. As is shown in Fig~\ref{5-1}, it applies temporal-ensemble techniques such as smoothing or temporal averaging to reduce per-step variance. This reduces error accumulation and improves effective control throughput, which benefits high-frequency operation. However, discontinuities or anomalous motions still occur at chunk boundaries. Real-Time Chunking (RTC)\cite{rtc} addresses this by reformulating chunk generation as a sequence inpainting problem. It freezes the already-executed prefix of each new chunk and applies a soft-masking scheme over the overlapping region to enforce cross-chunk consistency while still incorporating updated observations. Built on iterative denoising frameworks such as diffusion or flow matching \cite{flowmatching}, RTC provides smooth strategy transitions across chunks without sacrificing responsiveness, enabling robust real-time control.

Action chunking substantially increases per-inference action token length. In dual-arm tasks the sequence roughly doubles, which increases the load on autoregressive architectures. To address this, FAST\cite{fast} compresses discretized action sequences before inference. It applies a Discrete Cosine Transform (DCT) to move the sequence into the frequency domain, retains dominant low-frequency coefficients, and then applies Byte-Pair Encoding (BPE) to reduce token length. This lowers the cost during inference while allowing lossless reconstruction of the original actions, achieving significant computational savings. Similarly, OmniSAT\cite{omnisat} introduces a unified action tokenizer that converts continuous trajectories into compact discrete tokens. It first performs B-spline-based \cite{bspline} temporal alignment to obtain fixed-length control-point representations, then applies residual vector quantization (RVQ) \cite{rqv} to discretize them into multi-layer codebook indices, further grouped by positional, rotational, and gripper dimensions. This process yields a short, semantically structured token sequence that preserves trajectory fidelity while enabling efficient autoregressive modeling across diverse embodiments.

VOTE \cite{vote} introduces a single special token <ACT> that represents an entire action chunk, which is then mapped to continuous actions through a lightweight MLP. This design substantially reduces the number of generated tokens and decoding steps, enabling faster inference and lower training costs while maintaining strong task performance. During inference, VOTE further enhances robustness through an ensemble voting mechanism, where candidate actions from previous steps form a committee whose weighted votes determine the final output. SpatialVLA\cite{spatialvla} combines action discretization with action token compression, converting translational vectors into polar coordinates to decouple direction and distance, while retaining Euler angles for rotations and discretizing gripper states. 

In summary, action generation strategies evolve from directly predicting raw actions, to smoothing outputs through chunking, to compressing and restructuring action tokens. These methods complement each other in meeting high-frequency control demand, reducing latency, and mitigating error accumulation.

\subsection{Reasoning-Aware Action Generation}

The previous section focused on action-only VLA systems that generate raw actions directly. This approach offers a short inference pipeline, low latency, and a unified optimization objective for end-to-end training, which yields high computational efficiency and deployment simplicity. Direct action generation also maintains fluid execution in high-frequency control settings. However, it faces several limitations: limited interpretability, weaker cross-task generalization, and a substantial task gap with original VLM pretraining objectives, leading to underutilization of model potential.

\begin{figure}[htbp]
  \centering
  \includegraphics[width=0.85\textwidth]{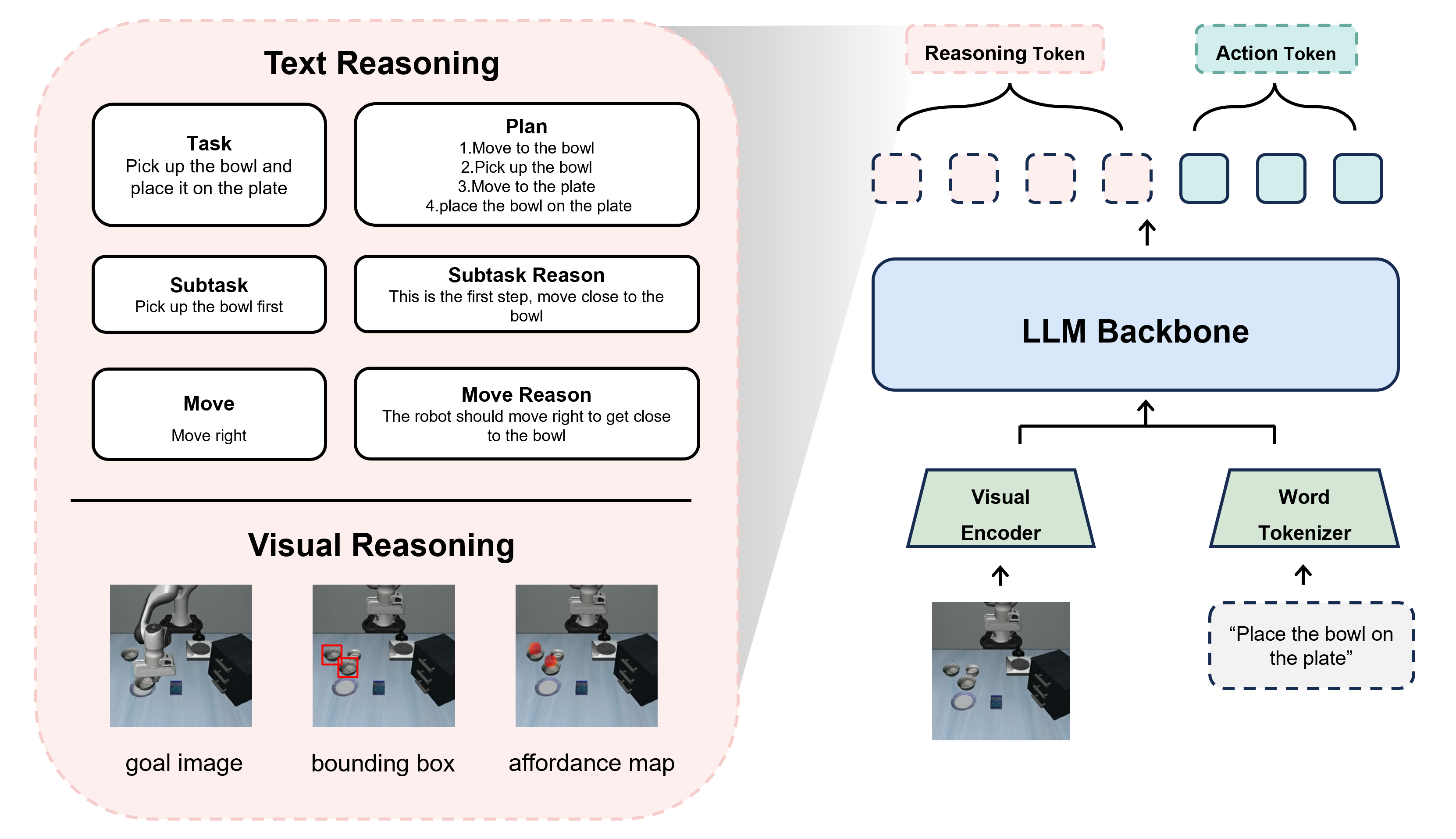} 
  \caption{VLA models augmented with reasoning. Beyond generating raw actions, these models produce intermediate reasoning outputs. Textual reasoning may include task lists, high-level plans, and detailed execution steps, while visual reasoning may involve goal states, bounding boxes, or affordances. Such reasoning enhances clarity in action generation.} 
  \label{5-2} 
\end{figure}

To address these shortcomings, an alternative paradigm explicitly incorporates reasoning stages before producing final actions. Typical reasoning steps include breaking down goals into subtasks, representing robot goals or motions with grounded features, and extracting task-relevant visuo-semantic features such as object bounding boxes or semantic keypoints. as illustrated in Figure~\ref{5-2}, these methods typically fall into language-based reasoning and vision-based reasoning approaches. By adding intermediate reasoning steps, they improve task decomposition and planning in long-horizon scenarios, and alleviates catastrophic forgetting in continual learning. However, reasoning introduces extra computational cost and latency, making efficiency-capability trade-offs a central challenge.

Language-based reasoning decomposes long, complex task instructions into fine-grained subtask descriptions for staged execution. Embodied Chain-of-Thought Reasoning (ECoT)\cite{ecot} applies chain-of-thought prompting to decompose instructions into structured fields (e.g., TASK, PLAN, SUBTASK, MOVE, GRIPPER, OBJECTS) before generating raw action sequences. While this provides clear task decomposition, it substantially increases sequence length (e.g., from 7 tokens per step in OpenVLA\cite{openvla} to 350 tokens per step in ECoT), thereby amplifying inference latency. To alleviate this overhead and make reasoning-based control more practical, subsequent variants aimed to streamline the reasoning process. Fast ECoT\cite{fastecot} further reduces runtime by reusing slowly-changing high-level plans and employing continuous batching to mitigate efficiency issues from variable-length sequences. In extensive experiments, ECoT-Lite\cite{ecotlite} trains to predict both reasoning and actions jointly but uses reasoning-dropout at test time, omitting reasoning tokens, outputs only actions to cut latency, which significantly boosts boost inference speed without a notable sacrifice in performance, thus achieving an effective trade-off between performance and efficiency.

Vision-based reasoning instead extracts spatial-semantic cues from visual input, such as affordances, keypoints, bounding boxes, or goal states, which guide action selection. UniPi\cite{unipi} trains a diffusion model for video prediction as a vision backbone, first forecasting a sequence of goal-state frames, then applying a pretrained inverse dynamics model to generate actions. This enables the model to learn from large-scale, action-free human video datasets. To reduce the heavy cost of full video generation, SuSIE\cite{susie} simplifies the output to a single subgoal image, which is then converted into actions via the inverse dynamics model, cutting inference delay. VPP\cite{vpp}voids full denoising by extracting multi-scale features from a pretrained Stable Video Diffusion model’s upsampling layers, aligning them in resolution, fusing via VideoFormer, and passing them to a diffusion policy for action generation. Although the single denoising step yields blurrier outputs, it captures future trajectory trends that are sufficient for planning. Most such methods adopt a two-stage pipeline—pretraining the video generation model separately and then combine it with the inverse dynamics model, which can limit cross-scene generalization. CoT-VLA\cite{cotvla} unifies subgoal image tokens and action tokens into a single autoregressive model, enabling end-to-end visual chain-of-thought reasoning. While this increases coupling between perception and control, it requires predicting ~256 image tokens in addition to actions, which slows inference by about 7×. DreamVLA\cite{dreamvla} addresses this with dynamic region-based forecasting, using optical flow to detect image regions relevant to end-effector or object motion, and predicting only these dynamic regions, preserving task-relevant information while greatly reducing computation. 
The summary of different types of reasoning-based VLA strategies is presented in the Table~\ref{tab:reasoning_methods}.

\begin{table}[h]
\centering
\scriptsize
\renewcommand{\arraystretch}{1.8}
\caption{Representative language-based and vision-based reasoning strategies in Vision-Language-Action models.}
\label{tab:reasoning_methods}
\begin{tabularx}{0.97\textwidth}{
    >{\raggedright\arraybackslash}m{1.6cm} 
    >{\raggedright\arraybackslash}m{1.6cm}  
    >{\raggedright\arraybackslash}m{2.2cm}  
    >{\raggedright\arraybackslash}m{4.0cm} 
    >{\raggedright\arraybackslash}m{4.7cm} 
}
\toprule
\textbf{~~~~~Modality} & \textbf{Method} & \textbf{Reasoning Output} & \textbf{~~~~~~~~~~~~~~~~~~Core Idea} & \textbf{~~~~~~~~~~~~~~~~~~Efficiency Strategy} \\
\midrule
\multirow{3}{*}{\makecell[cc]{\addlinespace[15pt]Language-based}}
  & ECoT\cite{ecot}      & High-level plans and actions & Decomposes instructions into structured task fields through CoT reasoning & Reuses parts of the reasoning chainacross multiple steps \\
  & Fast ECoT\cite{fastecot} & High-level plans and actions  & Follows the ECoT setting while optimizing reasoning generation speed & Reuses high-level reasoning and parallelizes modular reasoning steps across timesteps \\
  & ECoT-Lite\cite{ecotlite} & Actions only                  & Empirically studies how reasoning CoT affects success-speed trade-offs & Applies reasoning dropout to reduce inference latency \\
\midrule
\multirow{5}{*}{\makecell[cc]{\addlinespace[20pt]\\Vision-based}} 
  & UniPi\cite{unipi}     & Goal-state video frames & Diffusion-based video prediction and inverse dynamics for action inference & No explicit efficiency-oriented design, focusing on accurate visual trajectory generation \\
  & SuSIE\cite{susie}     & Single-frame subgoal image           & Predicts a single subgoal frame instead of a full trajectory & Reduces the cost by directly predicting intermediate subgoal frames. \\
  & VPP\cite{vpp}       & Multi-scale deffusion features    & Extracts and fuses intermediate features from pretrained video diffusion models & Skips full denoising by leveraging upsampling stage features \\
  & CoT-VLA\cite{cotvla}   & Goal image tokens and actions & End-to-end autoregressive reasoning for unified perception and control & Unifies reasoning across training and inference without redundant stages \\
  & DreamVLA\cite{dreamvla}  & Dynamic image regions         & Predicts only motion-relevant visual regions guided by optical flow & Selects motion-relevant regions to reduce visual computation \\
\bottomrule
\end{tabularx}
\end{table}

\subsection{Summary and Analysis}

In this section, we reviewed two representative paradigms for efficient output generation in VLA systems: direct raw action generation and reasoning-based action generation. The former emphasizes conceptual simplicity, direct compatibility with robotic control interfaces, and shorter output sequences, making it particularly suitable for real-time decision-making under strict frequency requirements. The latter, by contrast, incorporates explicit task planning and intermediate semantics, thereby enhancing long-horizon reasoning and cross-scene generalization. Language-based reasoning offers interpretability through human-readable plans, while vision-based reasoning conveys spatial semantics directly from perception to action, often allowing more parallel processing. Together, these two paradigms delineate the main pathways by which VLA systems aim to balance efficiency, generalization, and interpretability in action generation.

Despite these advances, both approaches present notable limitations. Direct action generation, while efficient in low-latency settings, struggles to transfer across different embodiments and often fails to sustain performance in long-horizon tasks, due to the absence of explicit reasoning signals. Reasoning-based methods, on the other hand, substantially increase computational burden by requiring the generation of long sequences of intermediate tokens. This extension of sequence length not only delays inference in high-frequency scenarios but also poses challenges for deployment in real-time control settings. Current solutions such as asynchronous reasoning and partial-step updates alleviate this problem to some extent, yet they remain ad-hoc and fall short of addressing the structural inefficiency inherent in reasoning-intensive paradigms.

Reasoning-based VLAs are likely to become the mainstream paradigm—either explicitly or implicitly—as reasoning greatly enhances generalization across diverse scenarios and offers interpretability crucial for real-world decision-making. Looking ahead, future research should focus on developing practical acceleration mechanisms that enable efficient reasoning without sacrificing these advantages. A central challenge is to reconcile the depth and flexibility of reasoning with the constraints of computational efficiency. Promising directions include selective reasoning, hierarchical planning, and hybrid architectures that balance interpretability, generalization, and speed, paving the way toward scalable and deployable reasoning-driven VLA agents.

\section {Efficient Training and Inference}

In the previous sections, we reviewed efficiency strategies from three perspectives: model architecture design, perception representation, and action generation. Yet structural compression and representation optimization alone are not enough for large-scale deployment. In real applications, the main bottlenecks come from computational resources such as FLOPs, memory, and inference latency, along with hardware limits. The efficiency of VLA systems also depends on how training can be streamlined and how inference can be accelerated. Based on this, the current section turns to efficiency-oriented methods for training and inference, focusing on knowledge distillation, quantization, parallel decoding, and other mechanisms that determine whether VLA models can be deployed in practice.

\subsection{Training Efficiency Techniques}

Training and deployment of VLA models pose substantial challenges due to the high cost of model adaptation and the limitations of hardware resources. Beyond optimizing inference, it is therefore essential to consider a broader optimization pipeline that spans the entire lifecycle of VLA models. Recent research has developed complementary strategies, parameter-efficient fine-tuning, knowledge distillation, parameter pruning and quantization, that together reduce resource consumption while maintaining competitive task performance.

Parameter-efficient fine-tuning (PEFT) is one effective approach. Tiny-VLA\cite{tinyvla} shows that expensive full-parameter adaptation of large pre-trained models is unnecessary. Methods such as Low-Rank Adaptation (LoRA)\cite{lora} freeze the original weights and insert a few trainable low-rank matrices to approximate the updates. This reduces the number of trainable parameters and memory usage by orders of magnitude. As a result, models can be adapted with only small-scale, domain-specific datasets and even deployed on edge devices or in data-constrained environments.

Knowledge distillation further improves efficiency by transferring performance from a larger model to a smaller one. Although distillation itself does not reduce training cost, it allows the distilled model to achieve nearly the same capability as the teacher while using far fewer parameters. For example, CEED-VLA\cite{ceedvla} uses consistency distillation to stabilize non-autoregressive inference. MoLE-VLA\cite{molevla} applies self-distillation to pass on semantic and control knowledge more effectively than direct pruning, preserving performance in the smaller model. VITA-VLA\cite{vitavla} formulates distillation as cross-model action alignment: it transfers a pretrained small action model’s motor control expertise into a large vision-language backbone (VITA-1.5-7B \cite{vita}). The process begins with lightweight alignment, where hidden representations of learnable action tokens in the VLA are matched to those of the expert model via MSE loss. This enables reuse of the expert’s pretrained action decoder. A subsequent end-to-end fine-tuning stage further refines multimodal fusion under supervision of action and gripper losses. Through this two-stage distillation, VITA-VLA effectively integrates visual-linguistic reasoning and precise control with minimal additional parameters.

Structured parameter pruning can reduce memory and computation in VLA models, but naive pruning often causes drastic drops in task success and safety. This is because VLA weights distribute important information across many directions, so removing even small-magnitude weights can discard critical subspaces. GLUESTICK \cite{gluestick} addresses this by a post-hoc, training-free recovery: it computes the difference between the original and pruned weights, extracts the dominant directions from this difference, and injects them back into the pruned model during inference. This low-rank correction restores the most important information lost to pruning while preserving efficiency, with a tunable trade-off between memory overhead and recovery fidelity. 

Quantization is crucial for deploying VLAs on resource-limited hardware but poses distinct challenges. Direct post-training quantization (PTQ), especially at low bit widths, often causes severe deviations in action outputs that can cascade into full task failures. To mitigate this, SQIL \cite{sqil} introduces a quantization-aware training (QAT) approach guided by a state-importance score (SIS), which measures how sensitive the policy is to perturbations in each state. During training, states with higher SIS values receive greater distillation weights, allowing the quantized model to preserve precision in critical decision states while maintaining overall efficiency. Building on this direction, BitVLA \cite{bitvla} further enhances quantization efficiency by applying ultra-low-bit QAT to the backbone. Through progressive distillation across BitNet \cite{bitnet} and SigLIP \cite{siglip}, it achieves substantial compression, reducing memory usage from about 15.1 GB in OpenVLA to roughly 1.4 GB, while retaining strong performance.

\subsection{Inference Efficiency Techniques}

\begin{figure}[htbp]
  \centering
  \includegraphics[width=0.65\textwidth]{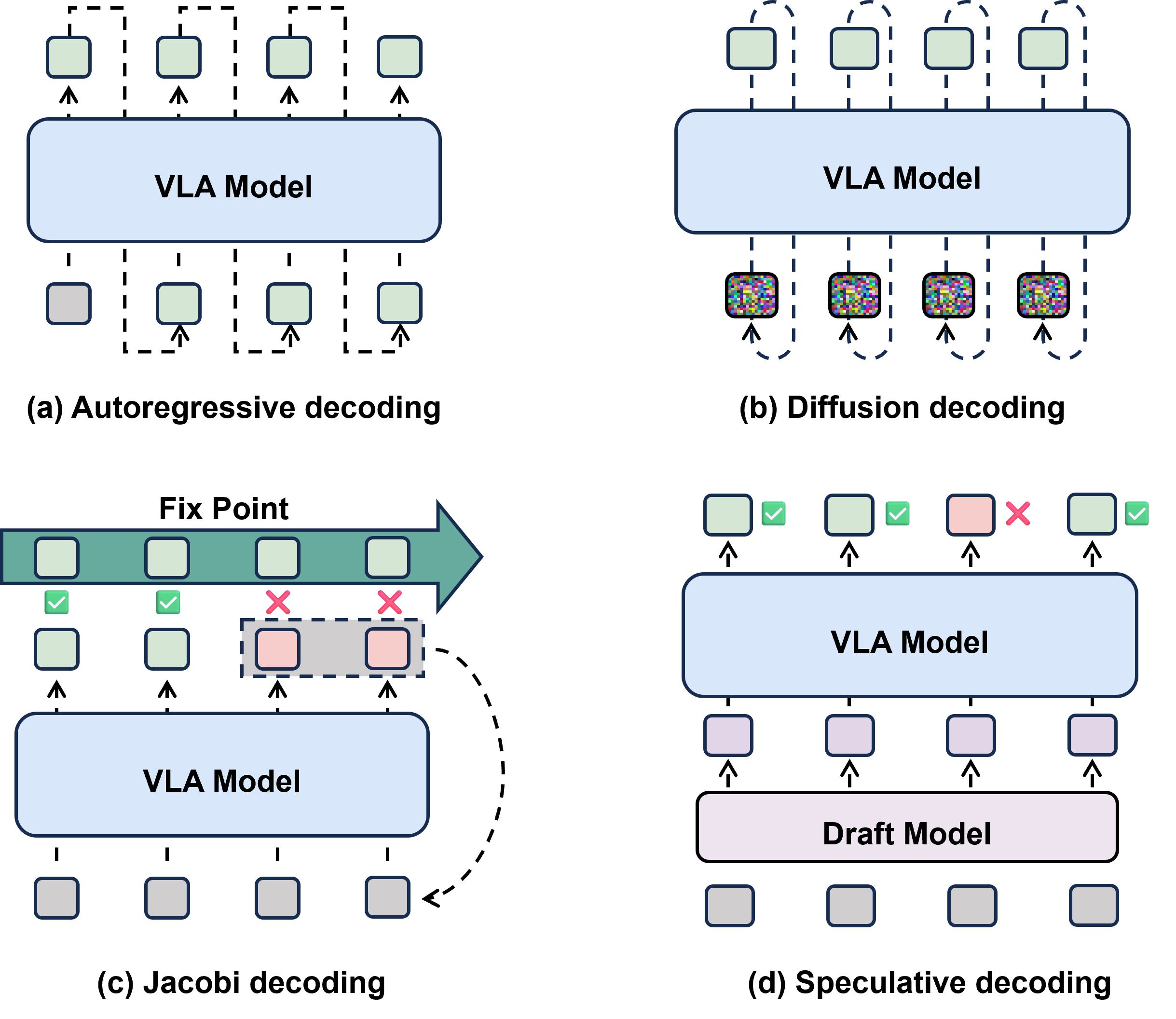} 
  \caption{Encoding mechanisms in VLA models. Subfigures (a) and (b) depict mainstream approaches based on autoregression and diffusion denoising, respectively. Subfigures (c) and (d) illustrate more recent designs for improving inference efficiency, including Jacobi decoding and speculative decoding.} 
  \label{6-1} 
\end{figure}

Conventional VLA models typically employ an autoregressive (AR) decoding paradigm that generates output tokens sequentially. This approach is simple to implement and efficient to train, but the inherent sequential dependency introduces a major computational bottleneck. In latency-sensitive applications such as high-frequency human-robot interaction or real-time robotic control, the lack of parallelism severely restricts response speed. Beyond AR decoding, another emerging paradigm is diffusion-based decoding, which generates outputs via a multistep denoising process. However, diffusion methods too suffer from slow inference due to the many iterative denoising steps required, making them suboptimal for real-time use. To address this limitation, recent work explores non-autoregressive (NAR) or parallel decoding paradigms, as shown in Figure~\ref{6-1}. These methods aim to reduce inference latency by enabling parallel computation, while adopting specific training strategies to preserve performance.

A more fundamental line of research re-engineers the architecture and action representation to align better with parallel inference. For example, openvla-oft\cite{openvla-oft} replaces causal attention with bidirectional attention, allowing the model to leverage both past and future context during encoding. This architectural change enables the model to predict an entire action sequence in a single parallel step. In addition, it reformulates action representation from a sequence of discrete tokens into continuous regression, and shifts the training objective from cross-entropy over next tokens to an L1 regression loss.

Speculative decoding\cite{specdecoding} provides another effective acceleration path. The Spec-VLA\cite{specvla} framework adapts this paradigm to VLA systems by introducing a two-stage process. A lightweight draft model first generates a candidate action sequence in parallel. Then, the full main model verifies the draft in a single pass. This approach reframes expensive step-by-step generation into fast parallel drafting followed by one-shot verification. To balance speed and accuracy, Spec-VLA relaxes the verification criterion, accepting drafts that are behaviorally valid even if they differ from the exact output of the main model.

Parallel decoding can also be realized through iterative refinement. Approaches such as PD-VLA\cite{pd-vla} draw inspiration from Jacobi iteration \cite{jacobi}, predicting all actions in parallel and refining them over multiple iterations until convergence. While this method offers high theoretical parallelism, it suffers from two challenges. First, a train-inference gap arises because training usually relies on standard autoregressive supervision, which does not expose the model to noisy contexts created during parallel decoding. Second, convergence may require many iterations in complex scenarios, reducing efficiency gains.CEED-VLA\cite{ceedvla} addresses these challenges with consistency distillation. During training, it uses the trajectory generated by a high-quality teacher model under Jacobi decoding as supervision, enabling the student model to learn to self-correct from imperfect states and converge reliably. An auxiliary autoregressive loss further ensures that the student’s output distribution stays aligned with that of the teacher. To improve efficiency, CEED-VLA also introduces an early-exit mechanism, which stops refinement once updates fall below a predefined threshold. This allows the model to finish decoding in only a few iterations for simpler tasks.
 
\subsection{Summary and Analysis}

This section has outlined recent progress in improving efficiency during both training and inference. On the training side, parameter-efficient tuning, knowledge distillation, and quantization reduce computational cost while keeping performance competitive. On the inference side, research is moving beyond pure autoregressive decoding and exploring parallel or hybrid generation schemes to speed up decision-making.

Despite these advances, important gaps remain. Most VLA frameworks are adapted from multimodal language models, and their efficiency methods often follow assumptions from general vision-language research. They rarely address robotics-specific needs such as temporal consistency, embodiment constraints, or action execution latency. Future work should design efficiency techniques that directly target embodied decision-making, rather than treating VLA optimization as a secondary outcome of VLM studies.

\section {Future Prospects}

The preceding sections have systematically reviewed efforts to improve the efficiency of VLA models across multiple facets, including model architecture, perception feature, action generation, and training-inference paradigms. Despite substantial advances, current models remain constrained when confronted with the complexity, variability, and uncertainty of real-world embodied tasks. Key limitations persist in areas such as redundant data usage, high-dimensional sensory processing, fragmented action sequences, and inefficient learning strategies, all of which impose significant computational and deployment costs.

This section aims to provide a forward-looking perspective, identifying the principal factors that will shape the next generation of efficient VLA systems. Rather than cataloging specific techniques, we examine the trade-offs between capability and efficiency, highlighting where improvements can most effectively reduce computational demand while preserving or enhancing task performance. Our discussion is organized around five interrelated dimensions: the co-optimization of models and data, the development of efficient spatio-temporal perception, the design of compact and continuous action representations, the transition from imitation learning to resource-aware reinforcement adaptation, and the establishment of an efficiency-centered evaluation framework. Together, these dimensions define a roadmap for research that systematically balances capability and efficiency, guiding the field toward practical, generalizable, and resource-conscious embodied intelligence.

\subsection{Model and Data: from Scale-Centric Design to Co-Optimized Efficiency}

In the evolution of VLA models, the historical paradigm of "larger models trained on more data" is approaching its saturation point. Future progress will increasingly rely on the co-evolution between model and data, where the representational capacity of the model and the structural composition of its training data are jointly optimized. This shift recognizes that efficiency and generalization no longer hinge solely on parameter scaling, but on how effectively a model can exploit, filter, and transfer information across diverse sources of visual, linguistic, and embodied datasets.

However, this trend toward model–data co-design also exposes a new layer of efficiency bottlenecks. The central challenge is no longer data scarcity per se, but data inefficiency, an imbalance between the exponential growth of data volume and the limited computational or memory budgets of embodied agents. Current datasets form a hierarchical "data pyramid", spanning internet-scale corpora, simulation trajectories, and real-world robot interactions. While large-scale data provides rich semantic grounding, the uncurated inclusion of redundant or low-value samples introduces unnecessary computation, prolongs training time, and dilutes high-quality physical feedback. Furthermore, the Sim-to-Real gap compounds this inefficiency: excessive simulation data can obscure the limited but critical contribution of real-world interaction samples. As a result, the overall learning process suffers from diminishing returns, each additional unit of data yields progressively smaller gains relative to its computational cost.

To address these inefficiencies, future efficiency-oriented research should extend beyond model compression toward the joint optimization of model and data. Promising directions include: developing data-centric efficiency frameworks that quantify the marginal utility of different data modalities and sources; exploring selective or curriculum-based data pipelines that dynamically adjust sampling frequency across the data pyramid; investigating joint scaling laws\cite{scalinglaw} that model how performance scales not only with parameters and data size but also with data quality and diversity. Such approaches would enable VLA models to allocate their limited training and inference budgets more intelligently, achieving a Pareto-optimal balance between data volume, model complexity, and embodied task performance. Ultimately, this co-evolutionary perspective reframes efficiency not as a constraint but as a guiding principle for sustainable scaling in embodied intelligence.

\subsection{Spatio-Temporal Perception: from 2D Frames to Efficient 3D Representations}

Current VLA models are constrained by two fundamental perceptual limitations: a spatial limitation rooted in the predominance of 2D image inputs, and a temporal myopia stemming from the Markovian assumption \cite{markovian} that decisions rely solely on the current frame. As embodied tasks grow in spatial and temporal complexity, requiring reasoning about occlusion, volumetric relations, and long-horizon intent, these simplifications increasingly fail. The natural evolution is therefore toward richer, three-dimensional and temporally grounded perceptual systems: explicit or implicit 3D representations (point clouds, voxel grids, neural fields), multi-view fusion, and persistent memory structures that record and summarize interaction histories.

Moving to such spatio-temporal representations, however, raises a practical efficiency dilemma. The added spatial fidelity and temporal context multiply the number of perceptual tokens (3D points, multi-view patches, or frame-wise embeddings) fed into the downstream LLM backbone. In VLA architectures where LLM perform the multimodal reasoning and action decoding, this token explosion translates directly into computational increases in attention and feedforward costs, larger activation footprints during inference, and greater latency, all of which threaten real-time control. Moreover, unselective retention of past observations blurs signal-to-noise ratios: vast swathes of redundant or low-value sensory history consume compute without commensurate gains in task success.

Given these constraints, future research should prioritize principled strategies to preserve spatial and temporal understanding while containing the token burden. At the spatial level, work may investigate task-aware 3D summarization that encodes high-fidelity geometry only in regions relevant to imminent interaction and represents background context with compact implicit descriptors. Temporally, approaches that combine short-term dense state tracking with long-term sparse summaries (keyframes or episodic sketches) can maintain continuity without unbounded history growth. Across modalities, semantic-guided filtering, using language or task priors to gate perception, and learnable token pruning during fusion can remove redundant signals before they enter the costly reasoning stage. Finally, evaluations should explicitly report the downstream backbone cost (token counts, attention FLOPs, inference latency) alongside success rates, to reveal practical Pareto frontiers between representational richness and deployable efficiency.

\subsection{Action Generation: from Chunked Commands to Compact Continuous Control}

Action generation stands at the core of a VLA system's embodiment capability. Current models predominantly operate under a chunked-control paradigm, where the model outputs discrete segments of action chunks to guide robot motion. This design enables efficient low-frequency control suitable for pick-and-place tasks. However, as robotic systems advance toward high-precision and high-frequency operations, such as assembly or dexterous manipulation, the temporal granularity and dynamic continuity required for control will inevitably increase. The emerging trend is thus a shift from low-rate, discrete command generation toward continuous and context-aware control, where the model not only decides what to do but also how to modulate its actions in real time.

This evolution brings a new class of efficiency bottlenecks. Increasing action chunk length or control frequency directly expands the number of output tokens, shifting the computational burden from the perception side to the language backbone itself. The autoregressive decoding cost grows nonlinearly with sequence length, making real-time inference on embodied platforms increasingly challenging. Moreover, maintaining temporal consistency across chunks becomes difficult, because the lack of smooth transition between successive output segments often causes discontinuous motion or instability in execution. Attempts to embed explicit reasoning, such as Chain-of-Thought planning, further exacerbate latency and memory demands, creating an acute tension between deliberation depth and control responsiveness. In embodied settings where control cycles operate at tens or hundreds of hertz, such reasoning overhead is often untenable.

To reconcile these opposing demands, future research must explore compact and hierarchical representations of actions that balance expressiveness and efficiency. One promising direction is to organize control outputs into multi-level abstractions: low-level motion primitives handled reactively, and higher-level intentions encoded through latent or parameterized tokens, thereby compressing output bandwidth without sacrificing control fidelity. Another key avenue is developing mechanisms for cross-chunk temporal coherence, for instance, caching or reusing hidden states between consecutive action segments to ensure smooth dynamics while avoiding redundant computation. Finally, the integration of reasoning within VLA must evolve toward reactive reasoning: lightweight, on-demand internal planning that preserves interpretability without incurring prohibitive latency. Such designs point toward a future where VLA systems achieve a true synthesis of reasoning and control, thinking and acting as one coherent, efficient process.

\subsection{Learning Paradigms: from Imitation to Efficient Reinforcement Adaptation}

Imitation Learning (IL), typically instantiated as behavior cloning \cite{bc}, remains the dominant paradigm for training VLA models due to its simplicity, stability, and high sample efficiency. By leveraging expert demonstrations, IL allows models to learn robust visuomotor mappings without costly trial-and-error interaction. However, this paradigm fundamentally constrains the ceiling of performance: the learned policy cannot exceed the coverage and quality of the demonstration dataset. As VLA systems pursue greater autonomy, adaptability, and generalization, the field is gradually shifting toward hybrid paradigms that incorporate elements of Reinforcement Learning (RL), enabling models not only to imitate but also to explore, optimize, and refine their own behavior beyond expert priors.

Yet this transition introduces a new spectrum of efficiency challenges. RL in embodied environments is notoriously sample-inefficient, requiring millions of interactions to achieve convergence, and such extensive exploration is impractical or unsafe on physical robots. Furthermore, VLA models compound the difficulty due to their large multimodal backbones, which increase the cost of each policy update. Reward specification for complex, language-conditioned manipulation tasks remains an open bottleneck, often producing sparse or ambiguous learning signals. Sim-to-real transfer further amplifies inefficiency: even policies refined in high-fidelity simulators tend to degrade in real-world deployment due to mismatched dynamics and sensing noise. Thus, while RL holds the promise of autonomy, its naive application to embodied agents risks becoming prohibitively expensive in both data and computation.

A promising research direction lies in redefining RL through the lens of efficiency. Rather than replacing imitation learning, future systems may employ progressive training pipelines that combine the strengths of both paradigms. An efficient learning hierarchy could begin with imitation learning to initialize a stable policy, followed by large-scale offline reinforcement fine-tuning using pre-collected interaction data to enhance generalization, and finally a limited stage of safety-constrained online adaptation in real environments. Complementary efficiency techniques, such as model-based rollouts, experience replay across multimodal contexts, and adaptive reward shaping, may further improve sample utilization and reduce hardware wear. In this view, the future of VLA training lies not in maximizing exploration, but in optimally allocating interaction budgets, achieving autonomous learning under strict efficiency constraints.

\subsection{Evaluation Criteria: from Fragmented Measures to Efficiency-Centric Benchmarking}

A shared and standardized evaluation framework is essential for meaningful progress in VLA research, particularly when efficiency is a central concern. Current VLA studies report heterogeneous metrics, rely on diverse datasets, and employ varied hardware, making cross-comparison difficult and obscuring true efficiency gains. Without a rigorous, reproducible benchmark, it is nearly impossible to quantify whether proposed techniques actually reduce computational cost without compromising task performance.

We argue that a structured, open benchmarking ecosystem can directly drive efficiency-oriented research. A three-dimensional evaluation framework, centered on resource efficiency, task performance, and interpretability, can guide the design and comparison of models with a dual focus on capability and computational cost:

\textbf{Resource and Efficiency:} Standardize reporting of model size, parameter count, inference latency, and hardware setup, alongside training time, memory footprint, and energy consumption. Such transparency allows equitable comparisons across different scales and platforms, and highlights the trade-offs between computational demand and task performance.
    
\textbf{Performance and Robustness:} Beyond task success rates, evaluation should consider long-horizon stability, recovery from failure, and resilience to environmental variations or sensor noise. Testing under distribution shifts and real-world perturbations ensures that performance metrics reflect true embodied competence rather than overfitting to narrow benchmarks.
    
\textbf{Interpretability and Auditability:} As embodied systems increasingly share control spaces with humans, understanding why a model acts is as important as what it does. Evaluation should therefore incorporate human-judged interpretability, visualization of action rationales, decision attribution mechanisms, and transparent action logging for post-hoc analysis—foundations for safety, accountability, and trust.
    
We hope the community will collaboratively establish a large-scale, multi-scenario, and multi-task open benchmark for VLA, similar to the role of ImageNet \cite{imagenet} in computer vision and GLUE \cite{glue} in natural language processing. Such a platform should provide open datasets, standardized simulation environments, and protocols for real-world robot testing. An open, transparent, and reproducible research ecosystem is necessary to measure progress and drive meaningful advances toward efficient embodied intelligence.

\section {Conclusion}

This survey reviews research on efficiency optimization in VLA models. We examine the progression from foundational model architectures, through perceptual representations, to high-level action generation, encompassing both training and inference. Building on this structure, we highlight several emerging directions that extend efficiency-focused research: the co-evolution of models and data, spatio-temporal perception to build dynamic world models, deliberative reasoning for intelligent action generation, learning paradigms that balance imitation and reinforcement strategies, and unified evaluation frameworks for reproducible assessment. Together, these directions show how efficiency improvements can enhance VLA systems as a whole. We hope this survey serves as a practical reference and supports the development of VLA systems that are both efficient and capable of general, reliable embodied intelligence.

\begin{thebibliography}{100}

\bibitem{llmsurvey}
Xunyu Zhu, Jian Li, Yong Liu, Can Ma, and Weiping Wang.
\newblock A survey on model compression for large language models.
\newblock {\em Transactions of the Association for Computational Linguistics},
  12:1556--1577, 2024.

\bibitem{ma1778survey}
Yueen Ma, Zixing Song, Yuzheng Zhuang, Jianye Hao, and Irwin King.
\newblock A survey on vision-language-action models for embodied ai (2024).
\newblock {\em arXiv preprint arXiv:2405.14093}, 1778.

\bibitem{zhong2025survey}
Yifan Zhong, Fengshuo Bai, Shaofei Cai, Xuchuan Huang, Zhang Chen, Xiaowei
  Zhang, Yuanfei Wang, Shaoyang Guo, Tianrui Guan, Ka~Nam Lui, et~al.
\newblock A survey on vision-language-action models: An action tokenization
  perspective.
\newblock {\em arXiv preprint arXiv:2507.01925}, 2025.

\bibitem{jiang2025survey}
Sicong Jiang, Zilin Huang, Kangan Qian, Ziang Luo, Tianze Zhu, Yang Zhong,
  Yihong Tang, Menglin Kong, Yunlong Wang, Siwen Jiao, et~al.
\newblock A survey on vision-language-action models for autonomous driving.
\newblock {\em arXiv preprint arXiv:2506.24044}, 2025.

\bibitem{rt1}
Anthony Brohan, Noah Brown, Justice Carbajal, Yevgen Chebotar, Joseph Dabis,
  Chelsea Finn, Keerthana Gopalakrishnan, Karol Hausman, Alex Herzog, Jasmine
  Hsu, et~al.
\newblock Rt-1: Robotics transformer for real-world control at scale.
\newblock {\em arXiv preprint arXiv:2212.06817}, 2022.

\bibitem{diffusionpolicy}
Cheng Chi, Zhenjia Xu, Siyuan Feng, Eric Cousineau, Yilun Du, Benjamin
  Burchfiel, Russ Tedrake, and Shuran Song.
\newblock Diffusion policy: Visuomotor policy learning via action diffusion.
\newblock {\em The International Journal of Robotics Research}, page
  02783649241273668, 2023.

\bibitem{rt2}
Brianna Zitkovich, Tianhe Yu, Sichun Xu, Peng Xu, Ted Xiao, Fei Xia, Jialin Wu,
  Paul Wohlhart, Stefan Welker, Ayzaan Wahid, et~al.
\newblock Rt-2: Vision-language-action models transfer web knowledge to robotic
  control.
\newblock In {\em Conference on Robot Learning}, pages 2165--2183. PMLR, 2023.

\bibitem{openvla}
Moo~Jin Kim, Karl Pertsch, Siddharth Karamcheti, Ted Xiao, Ashwin Balakrishna,
  Suraj Nair, Rafael Rafailov, Ethan Foster, Grace Lam, Pannag Sanketi, et~al.
\newblock Openvla: An open-source vision-language-action model.
\newblock {\em arXiv preprint arXiv:2406.09246}, 2024.

\bibitem{siglip}
Michael Tschannen, Alexey Gritsenko, Xiao Wang, Muhammad~Ferjad Naeem, Ibrahim
  Alabdulmohsin, Nikhil Parthasarathy, Talfan Evans, Lucas Beyer, Ye~Xia, Basil
  Mustafa, et~al.
\newblock Siglip 2: Multilingual vision-language encoders with improved
  semantic understanding, localization, and dense features.
\newblock {\em arXiv preprint arXiv:2502.14786}, 2025.

\bibitem{dinov2}
Maxime Oquab, Timoth{\'e}e Darcet, Th{\'e}o Moutakanni, Huy Vo, Marc
  Szafraniec, Vasil Khalidov, Pierre Fernandez, Daniel Haziza, Francisco Massa,
  Alaaeldin El-Nouby, et~al.
\newblock Dinov2: Learning robust visual features without supervision.
\newblock {\em arXiv preprint arXiv:2304.07193}, 2023.

\bibitem{llama}
Hugo Touvron, Louis Martin, Kevin Stone, Peter Albert, Amjad Almahairi, Yasmine
  Babaei, Nikolay Bashlykov, Soumya Batra, Prajjwal Bhargava, Shruti Bhosale,
  et~al.
\newblock Llama 2: Open foundation and fine-tuned chat models.
\newblock {\em arXiv preprint arXiv:2307.09288}, 2023.

\bibitem{octo}
Octo~Model Team, Dibya Ghosh, Homer Walke, Karl Pertsch, Kevin Black, Oier
  Mees, Sudeep Dasari, Joey Hejna, Tobias Kreiman, Charles Xu, et~al.
\newblock Octo: An open-source generalist robot policy.
\newblock {\em arXiv preprint arXiv:2405.12213}, 2024.

\bibitem{gr1}
Hongtao Wu, Ya~Jing, Chilam Cheang, Guangzeng Chen, Jiafeng Xu, Xinghang Li,
  Minghuan Liu, Hang Li, and Tao Kong.
\newblock Unleashing large-scale video generative pre-training for visual robot
  manipulation.
\newblock {\em arXiv preprint arXiv:2312.13139}, 2023.

\bibitem{gr2}
Chi-Lam Cheang, Guangzeng Chen, Ya~Jing, Tao Kong, Hang Li, Yifeng Li, Yuxiao
  Liu, Hongtao Wu, Jiafeng Xu, Yichu Yang, et~al.
\newblock Gr-2: A generative video-language-action model with web-scale
  knowledge for robot manipulation.
\newblock {\em arXiv preprint arXiv:2410.06158}, 2024.

\bibitem{gr3}
Chilam Cheang, Sijin Chen, Zhongren Cui, Yingdong Hu, Liqun Huang, Tao Kong,
  Hang Li, Yifeng Li, Yuxiao Liu, Xiao Ma, et~al.
\newblock Gr-3 technical report.
\newblock {\em arXiv preprint arXiv:2507.15493}, 2025.

\bibitem{vpp}
Yucheng Hu, Yanjiang Guo, Pengchao Wang, Xiaoyu Chen, Yen-Jen Wang, Jianke
  Zhang, Koushil Sreenath, Chaochao Lu, and Jianyu Chen.
\newblock Video prediction policy: A generalist robot policy with predictive
  visual representations.
\newblock {\em arXiv preprint arXiv:2412.14803}, 2024.

\bibitem{conrft}
Yuhui Chen, Shuai Tian, Shugao Liu, Yingting Zhou, Haoran Li, and Dongbin Zhao.
\newblock Conrft: A reinforced fine-tuning method for vla models via
  consistency policy.
\newblock {\em arXiv preprint arXiv:2502.05450}, 2025.

\bibitem{oxe}
Abby O’Neill, Abdul Rehman, Abhiram Maddukuri, Abhishek Gupta, Abhishek
  Padalkar, Abraham Lee, Acorn Pooley, Agrim Gupta, Ajay Mandlekar, Ajinkya
  Jain, et~al.
\newblock Open x-embodiment: Robotic learning datasets and rt-x models: Open
  x-embodiment collaboration 0.
\newblock In {\em 2024 IEEE International Conference on Robotics and Automation
  (ICRA)}, pages 6892--6903. IEEE, 2024.

\bibitem{droid}
Alexander Khazatsky, Karl Pertsch, Suraj Nair, Ashwin Balakrishna, Sudeep
  Dasari, Siddharth Karamcheti, Soroush Nasiriany, Mohan~Kumar Srirama,
  Lawrence~Yunliang Chen, Kirsty Ellis, et~al.
\newblock Droid: A large-scale in-the-wild robot manipulation dataset.
\newblock {\em arXiv preprint arXiv:2403.12945}, 2024.

\bibitem{pi0}
Kevin Black, Noah Brown, Danny Driess, Adnan Esmail, Michael Equi, Chelsea
  Finn, Niccolo Fusai, Lachy Groom, Karol Hausman, Brian Ichter, et~al.
\newblock $\pi$0: A vision-language-action flow model for general robot
  control. corr, abs/2410.24164, 2024. doi: 10.48550.
\newblock {\em arXiv preprint ARXIV.2410.24164}.

\bibitem{dit}
William Peebles and Saining Xie.
\newblock Scalable diffusion models with transformers.
\newblock In {\em Proceedings of the IEEE/CVF international conference on
  computer vision}, pages 4195--4205, 2023.

\bibitem{flowmatching}
Yaron Lipman, Ricky~TQ Chen, Heli Ben-Hamu, Maximilian Nickel, and Matt Le.
\newblock Flow matching for generative modeling.
\newblock {\em arXiv preprint arXiv:2210.02747}, 2022.

\bibitem{robomamba}
Jiaming Liu, Mengzhen Liu, Zhenyu Wang, Lily Lee, Kaichen Zhou, Pengju An,
  Senqiao Yang, Renrui Zhang, Yandong Guo, and Shanghang Zhang.
\newblock Robomamba: Multimodal state space model for efficient robot reasoning
  and manipulation.
\newblock {\em arXiv e-prints}, pages arXiv--2406, 2024.

\bibitem{mamba}
Albert Gu and Tri Dao.
\newblock Mamba: Linear-time sequence modeling with selective state spaces.
\newblock {\em arXiv preprint arXiv:2312.00752}, 2023.

\bibitem{tinyvla}
Junjie Wen, Yichen Zhu, Jinming Li, Minjie Zhu, Zhibin Tang, Kun Wu, Zhiyuan
  Xu, Ning Liu, Ran Cheng, Chaomin Shen, et~al.
\newblock Tinyvla: Towards fast, data-efficient vision-language-action models
  for robotic manipulation.
\newblock {\em IEEE Robotics and Automation Letters}, 2025.

\bibitem{pythia}
Stella Biderman, Hailey Schoelkopf, Quentin~Gregory Anthony, Herbie Bradley,
  Kyle O’Brien, Eric Hallahan, Mohammad~Aflah Khan, Shivanshu Purohit,
  USVSN~Sai Prashanth, Edward Raff, et~al.
\newblock Pythia: A suite for analyzing large language models across training
  and scaling.
\newblock In {\em International Conference on Machine Learning}, pages
  2397--2430. PMLR, 2023.

\bibitem{smolvla}
Mustafa Shukor, Dana Aubakirova, Francesco Capuano, Pepijn Kooijmans, Steven
  Palma, Adil Zouitine, Michel Aractingi, Caroline Pascal, Martino Russi,
  Andres Marafioti, et~al.
\newblock Smolvla: A vision-language-action model for affordable and efficient
  robotics.
\newblock {\em arXiv preprint arXiv:2506.01844}, 2025.

\bibitem{smolvlm}
Andr{\'e}s Marafioti, Orr Zohar, Miquel Farr{\'e}, Merve Noyan, Elie Bakouch,
  Pedro Cuenca, Cyril Zakka, Loubna~Ben Allal, Anton Lozhkov, Nouamane Tazi,
  et~al.
\newblock Smolvlm: Redefining small and efficient multimodal models.
\newblock {\em arXiv preprint arXiv:2504.05299}, 2025.

\bibitem{nora}
Chia-Yu Hung, Qi~Sun, Pengfei Hong, Amir Zadeh, Chuan Li, U~Tan, Navonil
  Majumder, Soujanya Poria, et~al.
\newblock Nora: A small open-sourced generalist vision language action model
  for embodied tasks.
\newblock {\em arXiv preprint arXiv:2504.19854}, 2025.

\bibitem{qwen}
Jinze Bai, Shuai Bai, Yunfei Chu, Zeyu Cui, Kai Dang, Xiaodong Deng, Yang Fan,
  Wenbin Ge, Yu~Han, Fei Huang, et~al.
\newblock Qwen technical report.
\newblock {\em arXiv preprint arXiv:2309.16609}, 2023.

\bibitem{flower}
Moritz Reuss, Hongyi Zhou, Marcel R{\"u}hle, {\"O}mer~Erdin{\c{c}}
  Ya{\u{g}}murlu, Fabian Otto, and Rudolf Lioutikov.
\newblock Flower: Democratizing generalist robot policies with efficient
  vision-language-action flow policies.
\newblock {\em arXiv preprint arXiv:2509.04996}, 2025.

\bibitem{deervla}
Yang Yue, Yulin Wang, Bingyi Kang, Yizeng Han, Shenzhi Wang, Shiji Song, Jiashi
  Feng, and Gao Huang.
\newblock Deer-vla: Dynamic inference of multimodal large language models for
  efficient robot execution.
\newblock {\em Advances in Neural Information Processing Systems},
  37:56619--56643, 2024.

\bibitem{molevla}
Rongyu Zhang, Menghang Dong, Yuan Zhang, Liang Heng, Xiaowei Chi, Gaole Dai,
  Li~Du, Yuan Du, and Shanghang Zhang.
\newblock Mole-vla: Dynamic layer-skipping vision language action model via
  mixture-of-layers for efficient robot manipulation.
\newblock {\em arXiv preprint arXiv:2503.20384}, 2025.

\bibitem{moe}
Robert~A Jacobs, Michael~I Jordan, Steven~J Nowlan, and Geoffrey~E Hinton.
\newblock Adaptive mixtures of local experts.
\newblock {\em Neural computation}, 3(1):79--87, 1991.

\bibitem{selfdistill}
Linfeng Zhang, Jiebo Song, Anni Gao, Jingwei Chen, Chenglong Bao, and Kaisheng
  Ma.
\newblock Be your own teacher: Improve the performance of convolutional neural
  networks via self distillation.
\newblock In {\em Proceedings of the IEEE/CVF international conference on
  computer vision}, pages 3713--3722, 2019.

\bibitem{efficientvla}
Yantai Yang, Yuhao Wang, Zichen Wen, Luo Zhongwei, Chang Zou, Zhipeng Zhang,
  Chuan Wen, and Linfeng Zhang.
\newblock Efficientvla: Training-free acceleration and compression for
  vision-language-action models.
\newblock {\em arXiv preprint arXiv:2506.10100}, 2025.

\bibitem{thinking}
Daniel Kahneman.
\newblock {\em Thinking, fast and slow}.
\newblock macmillan, 2011.

\bibitem{lcb}
Yide Shentu, Philipp Wu, Aravind Rajeswaran, and Pieter Abbeel.
\newblock From llms to actions: Latent codes as bridges in hierarchical robot
  control.
\newblock In {\em 2024 IEEE/RSJ International Conference on Intelligent Robots
  and Systems (IROS)}, pages 8539--8546. IEEE, 2024.

\bibitem{llava}
Haotian Liu, Chunyuan Li, Qingyang Wu, and Yong~Jae Lee.
\newblock Visual instruction tuning.
\newblock {\em Advances in neural information processing systems},
  36:34892--34916, 2023.

\bibitem{3ddiffueractor}
Tsung-Wei Ke, Nikolaos Gkanatsios, and Katerina Fragkiadaki.
\newblock 3d diffuser actor: Policy diffusion with 3d scene representations.
\newblock {\em arXiv preprint arXiv:2402.10885}, 2024.

\bibitem{hirt}
Jianke Zhang, Yanjiang Guo, Xiaoyu Chen, Yen-Jen Wang, Yucheng Hu, Chengming
  Shi, and Jianyu Chen.
\newblock Hirt: Enhancing robotic control with hierarchical robot transformers.
\newblock {\em arXiv preprint arXiv:2410.05273}, 2024.

\bibitem{instructblip}
Wenliang Dai, Junnan Li, Dongxu Li, Anthony Tiong, Junqi Zhao, Weisheng Wang,
  Boyang Li, Pascale~N Fung, and Steven Hoi.
\newblock Instructblip: Towards general-purpose vision-language models with
  instruction tuning.
\newblock {\em Advances in neural information processing systems},
  36:49250--49267, 2023.

\bibitem{efficientnet}
Brett Koonce.
\newblock Efficientnet.
\newblock In {\em Convolutional neural networks with swift for Tensorflow:
  image recognition and dataset categorization}, pages 109--123. Springer,
  2021.

\bibitem{map}
Juho Lee, Yoonho Lee, Jungtaek Kim, Adam Kosiorek, Seungjin Choi, and Yee~Whye
  Teh.
\newblock Set transformer: A framework for attention-based
  permutation-invariant neural networks.
\newblock In {\em International conference on machine learning}, pages
  3744--3753. PMLR, 2019.

\bibitem{robodual}
Qingwen Bu, Hongyang Li, Li~Chen, Jisong Cai, Jia Zeng, Heming Cui, Maoqing
  Yao, and Yu~Qiao.
\newblock Towards synergistic, generalized, and efficient dual-system for
  robotic manipulation.
\newblock {\em arXiv preprint arXiv:2410.08001}, 2024.

\bibitem{resampler}
Jean-Baptiste Alayrac, Jeff Donahue, Pauline Luc, Antoine Miech, Iain Barr,
  Yana Hasson, Karel Lenc, Arthur Mensch, Katherine Millican, Malcolm Reynolds,
  et~al.
\newblock Flamingo: a visual language model for few-shot learning.
\newblock {\em Advances in neural information processing systems},
  35:23716--23736, 2022.

\bibitem{openhelix}
Can Cui, Pengxiang Ding, Wenxuan Song, Shuanghao Bai, Xinyang Tong, Zirui Ge,
  Runze Suo, Wanqi Zhou, Yang Liu, Bofang Jia, et~al.
\newblock Openhelix: A short survey, empirical analysis, and open-source
  dual-system vla model for robotic manipulation.
\newblock {\em arXiv preprint arXiv:2505.03912}, 2025.

\bibitem{fis}
Hao Chen, Jiaming Liu, Chenyang Gu, Zhuoyang Liu, Renrui Zhang, Xiaoqi Li, Xiao
  He, Yandong Guo, Chi-Wing Fu, Shanghang Zhang, et~al.
\newblock Fast-in-slow: A dual-system foundation model unifying fast
  manipulation within slow reasoning.
\newblock {\em arXiv preprint arXiv:2506.01953}, 2025.

\bibitem{hume}
Haoming Song, Delin Qu, Yuanqi Yao, Qizhi Chen, Qi~Lv, Yiwen Tang, Modi Shi,
  Guanghui Ren, Maoqing Yao, Bin Zhao, et~al.
\newblock Hume: Introducing system-2 thinking in visual-language-action model.
\newblock {\em arXiv preprint arXiv:2505.21432}, 2025.

\bibitem{hypervla}
Zheng Xiong, Kang Li, Zilin Wang, Matthew Jackson, Jakob Foerster, and Shimon
  Whiteson.
\newblock Hypervla: Efficient inference in vision-language-action models via
  hypernetworks.
\newblock {\em arXiv preprint arXiv:2510.04898}, 2025.

\bibitem{spvla}
Ye~Li, Yuan Meng, Zewen Sun, Kangye Ji, Chen Tang, Jiajun Fan, Xinzhu Ma,
  Shutao Xia, Zhi Wang, and Wenwu Zhu.
\newblock Sp-vla: A joint model scheduling and token pruning approach for vla
  model acceleration.
\newblock {\em arXiv preprint arXiv:2506.12723}, 2025.

\bibitem{ridge}
Gary~C McDonald.
\newblock Ridge regression.
\newblock {\em Wiley Interdisciplinary Reviews: Computational Statistics},
  1(1):93--100, 2009.

\bibitem{rtcache}
Owen Kwon, Abraham George, Alison Bartsch, and Amir~Barati Farimani.
\newblock Rt-cache: Efficient robot trajectory retrieval system.
\newblock {\em arXiv preprint arXiv:2505.09040}, 2025.

\bibitem{fastv}
Liang Chen, Haozhe Zhao, Tianyu Liu, Shuai Bai, Junyang Lin, Chang Zhou, and
  Baobao Chang.
\newblock An image is worth 1/2 tokens after layer 2: Plug-and-play inference
  acceleration for large vision-language models.
\newblock In {\em European Conference on Computer Vision}, pages 19--35.
  Springer, 2024.

\bibitem{flashattention}
Tri Dao, Dan Fu, Stefano Ermon, Atri Rudra, and Christopher R{\'e}.
\newblock Flashattention: Fast and memory-efficient exact attention with
  io-awareness.
\newblock {\em Advances in neural information processing systems},
  35:16344--16359, 2022.

\bibitem{flashvla}
Xudong Tan, Yaoxin Yang, Peng Ye, Jialin Zheng, Bizhe Bai, Xinyi Wang, Jia Hao,
  and Tao Chen.
\newblock Think twice, act once: Token-aware compression and action reuse for
  efficient inference in vision-language-action models.
\newblock {\em arXiv preprint arXiv:2505.21200}, 2025.

\bibitem{lightvla}
Titong Jiang, Xuefeng Jiang, Yuan Ma, Xin Wen, Bailin Li, Kun Zhan, Peng Jia,
  Yahui Liu, Sheng Sun, and Xianpeng Lang.
\newblock The better you learn, the smarter you prune: Towards efficient
  vision-language-action models via differentiable token pruning.
\newblock {\em arXiv preprint arXiv:2509.12594}, 2025.

\bibitem{adp}
Xiaohuan Pei, Yuxing Chen, Siyu Xu, Yunke Wang, Yuheng Shi, and Chang Xu.
\newblock Action-aware dynamic pruning for efficient vision-language-action
  manipulation.
\newblock {\em arXiv preprint arXiv:2509.22093}, 2025.

\bibitem{fastdrivevla}
Jiajun Cao, Qizhe Zhang, Peidong Jia, Xuhui Zhao, Bo~Lan, Xiaoan Zhang, Xiaobao
  Wei, Sixiang Chen, Zhuo Li, Yang Wang, et~al.
\newblock Fastdrivevla: Efficient end-to-end driving via plug-and-play
  reconstruction-based token pruning.
\newblock {\em arXiv preprint arXiv:2507.23318}, 2025.

\bibitem{specprunevla}
Hanzhen Wang, Jiaming Xu, Jiayi Pan, Yongkang Zhou, and Guohao Dai.
\newblock Specprune-vla: Accelerating vision-language-action models via
  action-aware self-speculative pruning.
\newblock {\em arXiv preprint arXiv:2509.05614}, 2025.

\bibitem{sqapvla}
Hengyu Fang, Yijiang Liu, Yuan Du, Li~Du, and Huanrui Yang.
\newblock Sqap-vla: A synergistic quantization-aware pruning framework for
  high-performance vision-language-action models.
\newblock {\em arXiv preprint arXiv:2509.09090}, 2025.

\bibitem{otter}
Huang Huang, Fangchen Liu, Letian Fu, Tingfan Wu, Mustafa Mukadam, Jitendra
  Malik, Ken Goldberg, and Pieter Abbeel.
\newblock Otter: A vision-language-action model with text-aware visual feature
  extraction.
\newblock {\em arXiv preprint arXiv:2503.03734}, 2025.

\bibitem{univla}
Yuqi Wang, Xinghang Li, Wenxuan Wang, Junbo Zhang, Yingyan Li, Yuntao Chen,
  Xinlong Wang, and Zhaoxiang Zhang.
\newblock Unified vision-language-action model.
\newblock {\em arXiv preprint arXiv:2506.19850}, 2025.

\bibitem{vla-cache}
Siyu Xu, Yunke Wang, Chenghao Xia, Dihao Zhu, Tao Huang, and Chang Xu.
\newblock Vla-cache: Towards efficient vision-language-action model via
  adaptive token caching in robotic manipulation.
\newblock {\em arXiv preprint arXiv:2502.02175}, 2025.

\bibitem{ttfvla}
Chenghao Liu, Jiachen Zhang, Chengxuan Li, Zhimu Zhou, Shixin Wu, Songfang
  Huang, and Huiling Duan.
\newblock Ttf-vla: Temporal token fusion via pixel-attention integration for
  vision-language-action models.
\newblock {\em arXiv preprint arXiv:2508.19257}, 2025.

\bibitem{cot}
Jason Wei, Xuezhi Wang, Dale Schuurmans, Maarten Bosma, Fei Xia, Ed~Chi, Quoc~V
  Le, Denny Zhou, et~al.
\newblock Chain-of-thought prompting elicits reasoning in large language
  models.
\newblock {\em Advances in neural information processing systems},
  35:24824--24837, 2022.

\bibitem{ecot}
Micha{\l} Zawalski, William Chen, Karl Pertsch, Oier Mees, Chelsea Finn, and
  Sergey Levine.
\newblock Robotic control via embodied chain-of-thought reasoning.
\newblock {\em arXiv preprint arXiv:2407.08693}, 2024.

\bibitem{fastecot}
Zhekai Duan, Yuan Zhang, Shikai Geng, Gaowen Liu, Joschka Boedecker, and
  Chris~Xiaoxuan Lu.
\newblock Fast ecot: Efficient embodied chain-of-thought via thoughts reuse.
\newblock {\em arXiv preprint arXiv:2506.07639}, 2025.

\bibitem{act}
Tony~Z Zhao, Vikash Kumar, Sergey Levine, and Chelsea Finn.
\newblock Learning fine-grained bimanual manipulation with low-cost hardware.
\newblock {\em arXiv preprint arXiv:2304.13705}, 2023.

\bibitem{rtc}
Kevin Black, Manuel~Y Galliker, and Sergey Levine.
\newblock Real-time execution of action chunking flow policies.
\newblock {\em arXiv preprint arXiv:2506.07339}, 2025.

\bibitem{fast}
Karl Pertsch, Kyle Stachowicz, Brian Ichter, Danny Driess, Suraj Nair, Quan
  Vuong, Oier Mees, Chelsea Finn, and Sergey Levine.
\newblock Fast: Efficient action tokenization for vision-language-action
  models.
\newblock {\em arXiv preprint arXiv:2501.09747}, 2025.

\bibitem{omnisat}
Huaihai Lyu, Chaofan Chen, Senwei Xie, Pengwei Wang, Xiansheng Chen, Shanghang
  Zhang, and Changsheng Xu.
\newblock Omnisat: Compact action token, faster auto regression.
\newblock {\em arXiv preprint arXiv:2510.09667}, 2025.

\bibitem{bspline}
Hartmut Prautzsch, Wolfgang Boehm, and Marco Paluszny.
\newblock {\em B{\'e}zier and B-spline techniques}.
\newblock Springer Science \& Business Media, 2002.

\bibitem{rqv}
Seungjae Lee, Yibin Wang, Haritheja Etukuru, H~Jin Kim, Nur Muhammad~Mahi
  Shafiullah, and Lerrel Pinto.
\newblock Behavior generation with latent actions.
\newblock {\em arXiv preprint arXiv:2403.03181}, 2024.

\bibitem{vote}
Juyi Lin, Amir Taherin, Arash Akbari, Arman Akbari, Lei Lu, Guangyu Chen,
  Taskin Padir, Xiaomeng Yang, Weiwei Chen, Yiqian Li, et~al.
\newblock Vote: vision-language-action optimization with trajectory ensemble
  voting.
\newblock {\em arXiv preprint arXiv:2507.05116}, 2025.

\bibitem{spatialvla}
Delin Qu, Haoming Song, Qizhi Chen, Yuanqi Yao, Xinyi Ye, Yan Ding, Zhigang
  Wang, JiaYuan Gu, Bin Zhao, Dong Wang, et~al.
\newblock Spatialvla: Exploring spatial representations for
  visual-language-action model.
\newblock {\em arXiv preprint arXiv:2501.15830}, 2025.

\bibitem{ecotlite}
William Chen, Suneel Belkhale, Suvir Mirchandani, Oier Mees, Danny Driess, Karl
  Pertsch, and Sergey Levine.
\newblock Training strategies for efficient embodied reasoning.
\newblock {\em arXiv preprint arXiv:2505.08243}, 2025.

\bibitem{unipi}
Yilun Du, Sherry Yang, Bo~Dai, Hanjun Dai, Ofir Nachum, Josh Tenenbaum, Dale
  Schuurmans, and Pieter Abbeel.
\newblock Learning universal policies via text-guided video generation.
\newblock {\em Advances in neural information processing systems},
  36:9156--9172, 2023.

\bibitem{susie}
Nicoleta Preda, Fabian Suchanek, Wenjun Yuan, and Gerhard Weikum.
\newblock Susie: Search using services and information extraction.
\newblock In {\em 2013 IEEE 29th International Conference on Data Engineering
  (ICDE)}, pages 218--229. IEEE, 2013.

\bibitem{cotvla}
Qingqing Zhao, Yao Lu, Moo~Jin Kim, Zipeng Fu, Zhuoyang Zhang, Yecheng Wu,
  Zhaoshuo Li, Qianli Ma, Song Han, Chelsea Finn, et~al.
\newblock Cot-vla: Visual chain-of-thought reasoning for vision-language-action
  models.
\newblock In {\em Proceedings of the Computer Vision and Pattern Recognition
  Conference}, pages 1702--1713, 2025.

\bibitem{dreamvla}
Wenyao Zhang, Hongsi Liu, Zekun Qi, Yunnan Wang, Xinqiang Yu, Jiazhao Zhang,
  Runpei Dong, Jiawei He, He~Wang, Zhizheng Zhang, et~al.
\newblock Dreamvla: a vision-language-action model dreamed with comprehensive
  world knowledge.
\newblock {\em arXiv preprint arXiv:2507.04447}, 2025.

\bibitem{lora}
Edward~J Hu, Yelong Shen, Phillip Wallis, Zeyuan Allen-Zhu, Yuanzhi Li, Shean
  Wang, Lu~Wang, Weizhu Chen, et~al.
\newblock Lora: Low-rank adaptation of large language models.
\newblock {\em ICLR}, 1(2):3, 2022.

\bibitem{ceedvla}
Wenxuan Song, Jiayi Chen, Pengxiang Ding, Yuxin Huang, Han Zhao, Donglin Wang,
  and Haoang Li.
\newblock Ceed-vla: Consistency vision-language-action model with early-exit
  decoding.
\newblock {\em arXiv preprint arXiv:2506.13725}, 2025.

\bibitem{vitavla}
Shaoqi Dong, Chaoyou Fu, Haihan Gao, Yi-Fan Zhang, Chi Yan, Chu Wu, Xiaoyu Liu,
  Yunhang Shen, Jing Huo, Deqiang Jiang, et~al.
\newblock Vita-vla: Efficiently teaching vision-language models to act via
  action expert distillation.
\newblock {\em arXiv preprint arXiv:2510.09607}, 2025.

\bibitem{vita}
Chaoyou Fu, Haojia Lin, Xiong Wang, Yi-Fan Zhang, Yunhang Shen, Xiaoyu Liu,
  Haoyu Cao, Zuwei Long, Heting Gao, Ke~Li, et~al.
\newblock Vita-1.5: Towards gpt-4o level real-time vision and speech
  interaction.
\newblock {\em arXiv preprint arXiv:2501.01957}, 2025.

\bibitem{gluestick}
Jason Jabbour, Dong-Ki Kim, Max Smith, Jay Patrikar, Radhika Ghosal, Youhui
  Wang, Ali Agha, Vijay~Janapa Reddi, and Shayegan Omidshafiei.
\newblock Don't run with scissors: Pruning breaks vla models but they can be
  recovered.
\newblock {\em arXiv preprint arXiv:2510.08464}, 2025.

\bibitem{sqil}
Siddharth Reddy, Anca~D Dragan, and Sergey Levine.
\newblock Sqil: Imitation learning via reinforcement learning with sparse
  rewards.
\newblock {\em arXiv preprint arXiv:1905.11108}, 2019.

\bibitem{bitvla}
Hongyu Wang, Chuyan Xiong, Ruiping Wang, and Xilin Chen.
\newblock Bitvla: 1-bit vision-language-action models for robotics
  manipulation.
\newblock {\em arXiv preprint arXiv:2506.07530}, 2025.

\bibitem{bitnet}
Hongyu Wang, Shuming Ma, Li~Dong, Shaohan Huang, Huaijie Wang, Lingxiao Ma, Fan
  Yang, Ruiping Wang, Yi~Wu, and Furu Wei.
\newblock Bitnet: Scaling 1-bit transformers for large language models.
\newblock {\em arXiv preprint arXiv:2310.11453}, 2023.

\bibitem{openvla-oft}
Moo~Jin Kim, Chelsea Finn, and Percy Liang.
\newblock Fine-tuning vision-language-action models: Optimizing speed and
  success.
\newblock {\em arXiv preprint arXiv:2502.19645}, 2025.

\bibitem{specdecoding}
Yaniv Leviathan, Matan Kalman, and Yossi Matias.
\newblock Fast inference from transformers via speculative decoding.
\newblock In {\em International Conference on Machine Learning}, pages
  19274--19286. PMLR, 2023.

\bibitem{specvla}
Songsheng Wang, Rucheng Yu, Zhihang Yuan, Chao Yu, Feng Gao, Yu~Wang, and
  Derek~F Wong.
\newblock Spec-vla: speculative decoding for vision-language-action models with
  relaxed acceptance.
\newblock {\em arXiv preprint arXiv:2507.22424}, 2025.

\bibitem{pd-vla}
Wenxuan Song, Jiayi Chen, Pengxiang Ding, Han Zhao, Wei Zhao, Zhide Zhong,
  Zongyuan Ge, Jun Ma, and Haoang Li.
\newblock Accelerating vision-language-action model integrated with action
  chunking via parallel decoding.
\newblock {\em arXiv preprint arXiv:2503.02310}, 2025.

\bibitem{jacobi}
Andrea Santilli, Silvio Severino, Emilian Postolache, Valentino Maiorca,
  Michele Mancusi, Riccardo Marin, and Emanuele Rodol{\`a}.
\newblock Accelerating transformer inference for translation via parallel
  decoding.
\newblock {\em arXiv preprint arXiv:2305.10427}, 2023.

\bibitem{scalinglaw}
Jared Kaplan, Sam McCandlish, Tom Henighan, Tom~B Brown, Benjamin Chess, Rewon
  Child, Scott Gray, Alec Radford, Jeffrey Wu, and Dario Amodei.
\newblock Scaling laws for neural language models.
\newblock {\em arXiv preprint arXiv:2001.08361}, 2020.

\bibitem{markovian}
Richard Bellman.
\newblock A markovian decision process.
\newblock {\em Journal of mathematics and mechanics}, pages 679--684, 1957.

\bibitem{bc}
Dean~A Pomerleau.
\newblock Alvinn: An autonomous land vehicle in a neural network.
\newblock {\em Advances in neural information processing systems}, 1, 1988.

\bibitem{rlhf}
Paul~F Christiano, Jan Leike, Tom Brown, Miljan Martic, Shane Legg, and Dario
  Amodei.
\newblock Deep reinforcement learning from human preferences.
\newblock {\em Advances in neural information processing systems}, 30, 2017.

\bibitem{imagenet}
Jia Deng, Wei Dong, Richard Socher, Li-Jia Li, Kai Li, and Li~Fei-Fei.
\newblock Imagenet: A large-scale hierarchical image database.
\newblock In {\em 2009 IEEE conference on computer vision and pattern
  recognition}, pages 248--255. Ieee, 2009.

\bibitem{glue}
Alex Wang, Amanpreet Singh, Julian Michael, Felix Hill, Omer Levy, and Samuel~R
  Bowman.
\newblock Glue: A multi-task benchmark and analysis platform for natural
  language understanding.
\newblock {\em arXiv preprint arXiv:1804.07461}, 2018.

\end{thebibliography}

\end{document}